\documentclass[10pt,twocolumn,letterpaper]{article}

\usepackage{cvpr}              

\usepackage{booktabs}
\usepackage{multirow}
\usepackage{makecell}
\usepackage[table]{xcolor}
\usepackage{threeparttable}
\usepackage{array}
\usepackage{subcaption}

\definecolor{cvprblue}{rgb}{0.21,0.49,0.74}
\usepackage[pagebackref,breaklinks,colorlinks,allcolors=cvprblue]{hyperref}

\def\paperID{45049} 
\def\confName{CVPR}
\def\confYear{2026}

\title{iMontage: Unified, Versatile, Highly Dynamic Many-to-many Image Generation}

\author{
Zhoujie Fu$^{1,2}$,
Xianfang Zeng$^{2,\ddagger}$,
Jinghong Lan$^{2}$,
Haoling Xie$^{2}$,
Xinyao Liao$^{1,2}$,
Cheng Chen$^{1}$, \\
Junyi Chen$^{3}$,
Jiacheng Wei$^{1}$, 
Wei Cheng$^{2}$,
Shiyu Liu$^{2}$,
Yunuo Chen$^{2,3}$,
Gang Yu$^{\dagger,2}$
Guosheng Lin$^{\dagger,1}$\\[4pt]
$^{1}$Nanyang Technological University \quad
$^{2}$StepFun \quad
$^{3}$Shanghai Jiao Tong University\\[4pt]
\href{https://kr1sjfu.github.io/iMontage-web/}{https://kr1sjfu.github.io/iMontage-web/}
}

\begin{document}

\twocolumn[{%
\renewcommand\twocolumn[1][]{#1}%
\maketitle
\includegraphics[width=\textwidth,height=0.8\textheight,keepaspectratio]{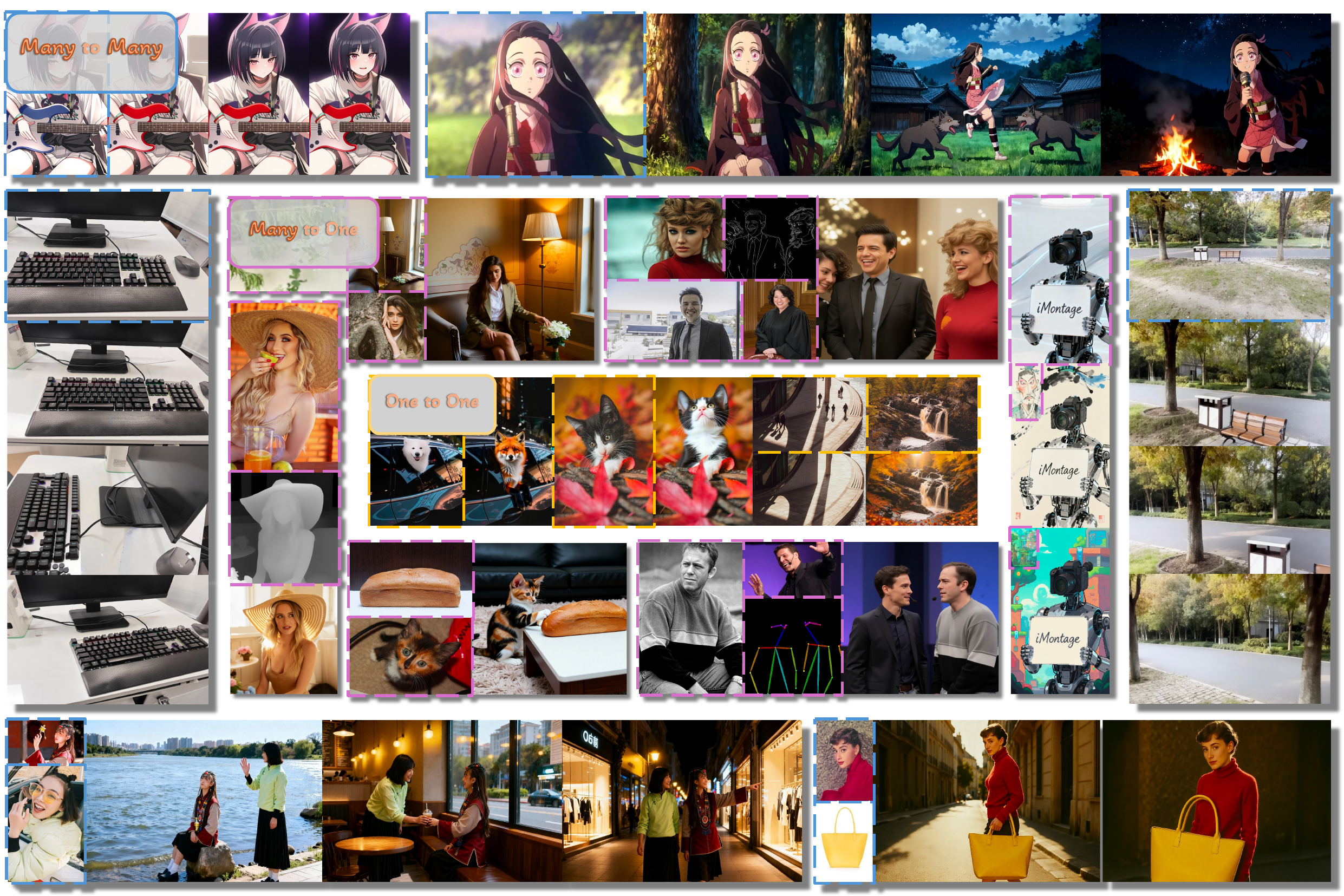}
\vspace{-10pt}
  \captionof{figure}{iMontage can flexibly deal with many input images, and can generate many output images with highly consistency. We use three different colors to represent three settings. The dotted-line box images are the input.}
\vspace{20pt}
\label{fig:teaser}
}]

\begin{abstract}
\let\thefootnote\relax\footnote{\textsuperscript{\ddag} Project Leader ~~\textsuperscript{\dag}~Corresponding Author}
Pre-trained video models learn powerful priors for generating high-quality, temporally coherent content. While these models excel at temporal coherence, their dynamics are often constrained by the continuous nature of their training data.  We hypothesize that by injecting the rich and unconstrained content diversity from image data into this coherent temporal framework, we can generate image sets that feature both natural transitions and a far more expansive dynamic range. To this end, we introduce iMontage, a unified framework designed to repurpose a powerful video model into an all-in-one image generator. The framework consumes and produces variable-length image sets, unifying a wide array of image generation and editing tasks. To achieve this, we propose an elegant and minimally invasive adaptation strategy, complemented by a tailored data curation process and training paradigm. This approach allows the model to acquire broad image manipulation capabilities without corrupting its invaluable original motion priors. iMontage excels across several mainstream many-in-many-out tasks, not only maintaining strong cross-image contextual consistency but also generating scenes with extraordinary dynamics that surpass conventional scopes. Our code and model weights will be made publicly available.
\vspace{-10pt}
\end{abstract}    
\section{Introduction}
\label{sec:intro}

Large-scale diffusion-based generative models\citep{rombach2022high, saharia2022photorealistic, podell2023sdxl, Flux.1, esser2024scaling, peebles2023scalable} have sparked a revolution in creative and high-quality image generation, accelerating progress in downstream tasks such as image editing. A recent trend in the field is to unify diverse image tasks within a single framework \citep{xiao2025omnigen, mao2025ace++, liu2025step1x, qin2025lumina}, inspired by the success of Large Language Models (LLMs) and large vision language models (VLMs) \citep{GPT4o, Gemini2.5, xie2024show, deng2025emerging}. While most unified image models remain specialized for single-in-single-out image tasks, certain commercial model has taken an early lead in extending unified image generation to multi-input, multi-output settings\citep{Seedream4.0} very recently. Accordingly, the many-to-many (multi-input, multi-output) setting warrants systematic exploration by the academic and open-source communities.

The many-to-many paradigm splits into two approaches: (i) token-centric models that represent text and images as a unified multimodal token stream and autoregressively generate target tokens conditioned on the inputs\citep{xiao2025omnigen}, thereby achieving many-to-many mappings; and (ii) video-centric pipelines that repurpose video diffusion generation as backbones, casting the task as discontinuous video generation and naturally accommodating variable numbers of input and output frames\citep{chen2025unireal, lin2025realgeneral}. While the first approach provides an appealing and promising modal-unified solution, it's generation quality and instruction following capability is challenged by common sense compared to diffusion paradigm. In contrast, the second approach elegantly leverages pre-trained motion priors to markedly enhance temporal coherence and handle variable-length inputs and outputs. Specifically, \citep{chen2025unireal} trained a model of video generation from scratch and constructed a large-scale dataset of captioned frame pairs for instruction-tuned editing, demonstrating strong consistency and faithful detail preservation with respect to the input images.

Despite these advances for the diffusion-based paradigm, a critical question persists: \textbf{How can a model generate highly dynamic multi-image outputs while maintaining temporal and semantic consistency?} To our empirical knowledge, image-only models can produce highly diverse images based on the same inputs, yet they struggle with temporal consistency due to limited implicit understanding of world dynamics. Meanwhile, video-based models bring strong motion priors that improve temporal consistency; however, most foundation video models are trained predominantly on contiguous clips, which rarely contains hard cuts, abrupt transitions, or large camera/subject motions, and thereby transferring poorly to highly dynamic content and limiting task versatility.

In response, we present \textbf{iMontage}, a unified generative model that produces multiple, highly dynamic images conditioned on instructions and arbitrary reference images. Following the video-based paradigm, iMontage builds on a large pretrained video model and treats both inputs and outputs as pseudo-frames. 
We introduce a novel rotary positional embedding (RoPE) strategy to prevent conceptual ambiguity between multiple image frames and video frames. 
Our strategy explicitly maintains the model's pre-trained capability in modeling temporal coherence, while clearly differentiating the discrete nature of image sets from the continuous flow of video sequences.
We further provide a data-curation pipeline, which is carefully categorized and filtered for motion diversity and instruction quality, supporting broad, highly dynamic scenario. 
Finally, we detail a training regimen that offers practical insights into multi-task unification. 
Together, these components marry video generation with many-to-many image generation, achieving both temporal and content consistency.

We evaluate iMontage across three settings: one-to-one image editing, many-to-one image generation, and many-to-many image generation. 
For each setting, we present strong qualitative performances across all sub-tasks, showcasing robust instruction following, high-dynamic outputs, and consistent content generation, as presented in \cref{fig:teaser}. Furthermore, we provide state-of-the art quantitative metrics on image editing benchmark (one-to-one), in-context learning benchmark (many-to-one) and storyboard generation evaluation (many-to-many).

In summary, our contributions are as follows:
\begin{itemize}
\item We introduce \textbf{iMontage}, a unified model that handles variable numbers of input and output frames, bridging video generation and highly dynamic image generation. 
\item We develop a \emph{task-agnostic, temporally diverse} data curation pipeline paired with a multi-task training paradigm, ensuring learnability across heterogeneous tasks and temporal structures and enabling robust many-to-many generalization. 
\item Our model showcases convincing results over huge number of variable experiments, including most mainstreaming image generation and editing tasks. Massive visualization results and comprehensive evaluation metrics provide SOTA results in open-source community and even comparable results with commercial models.
\end{itemize}

\begin{figure*}[t]
    \begin{center}
\includegraphics[width=\textwidth,height=0.9\textheight,keepaspectratio]{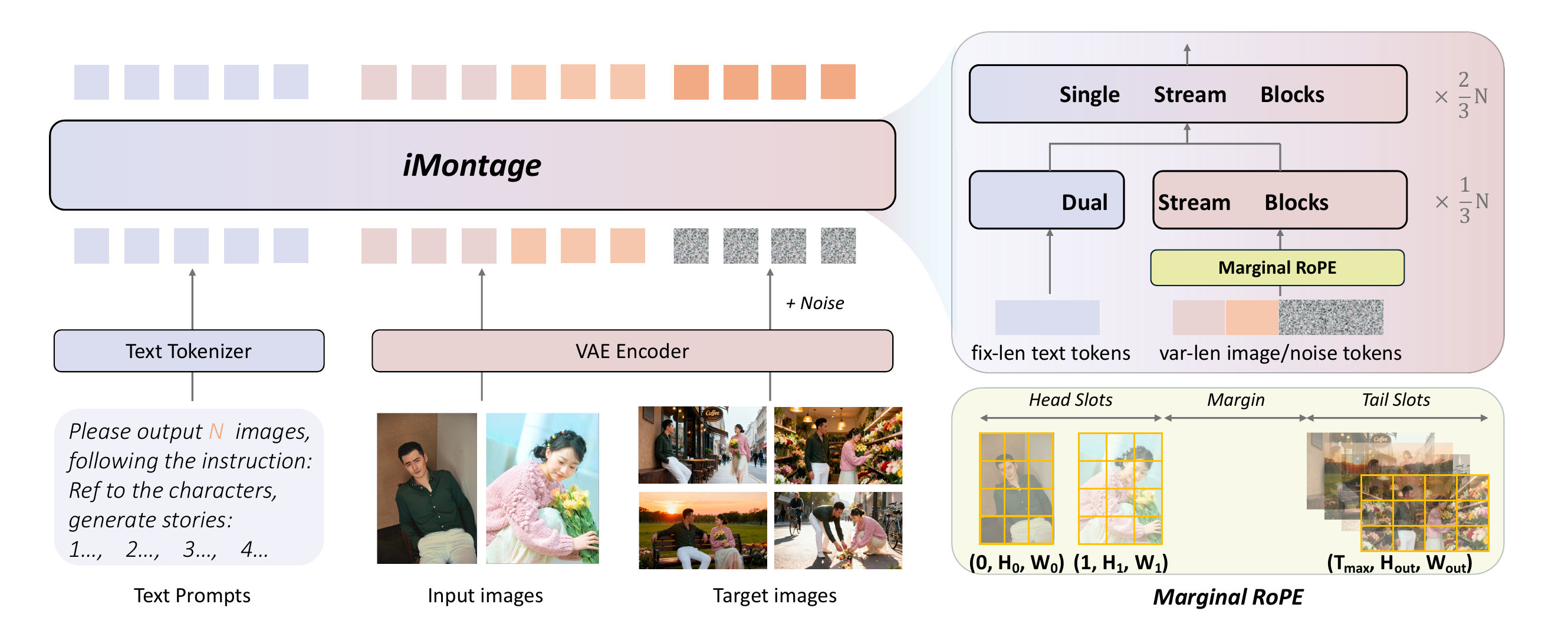}
\vspace{-30pt}
\end{center}
    \captionof{figure}{\textbf{Overview of iMontage.} The model accepts a flexible set of reference images and produces \textbf{N} outputs conditioned on a text prompt. Images are encoded by a 3D VAE separately, text by a language model, and both token streams are processed by an MMDiT. We concatenate clean reference-image tokens with noisy target tokens before denoising. \emph{Right:} training uses fixed-length text tokens and variable-length image/noise tokens, transitions from dual stream to single stream blocks. For image branch, we apply \emph{Marginal RoPE}, a head–tail temporal indexing that separates input and output pseudo-frames, preserves spatial RoPE, and supports many-to-many generation. In figure, notation H and W with subscription denote the height/width indices of the 2D RoPE computed at the image’s native resolution, while notation T represents assigned time index for temporal dimension.}
\vspace{-10pt}
\label{fig:arch}
\end{figure*}

\section{Related work}

\subsection{Unified Generation and Editing Models}

Recent research has increasingly focused on consolidating diverse visual synthesis tasks into single, unified frameworks. Early efforts\citep{xia2025dreamomni, fu2025univg, li2024unimo} like OmniGen \citep{xiao2025omnigen}  and ACE++ \citep{mao2025ace++} pioneered monolithic architectures capable of handling generation, editing, and other vision tasks without requiring task-specific modules. This trend evolved with the integration of powerful multimodal large language models (MLLMs) as reasoning engines. Models such as Step1X-Edit \citep{liu2025step1x} and Qwen-Image \citep{deng2025emerging}  leverage an MLLM to interpret complex user instructions, which then guide a diffusion decoder to produce high-fidelity edits. This approach significantly improves instruction-following capabilities. Notably, unified systems in other AIGC area are also emerging, such as \citep{bar2024lumiere, luo2025univid, wei2025univideo} in video generation, \citep{liu2024audioldm, yang2023uniaudio, vyas2023audiobox} in audio generation, and even more powerful combining different modalities together\citep{Sora2, Veo3, shan2025hunyuanvideo, team2025aether, chen2025deepverse}. While these unified image models demonstrate impressive versatility, they are predominantly architected for single-input, single-output tasks. They lack the inherent capability to manage multiple image inputs and generate a set of dynamically varied yet coherent outputs from a single prompt, a key limitation our work addresses.

\subsection{From One-to-one to Many-to-many Generation}

The frontier of generative modeling is advancing from single-image tasks to more complex many-to-many scenarios that require handling multiple inputs to produce multiple outputs. A significant paradigm shift was introduced by UniReal\citep{chen2025unireal} , which re-frames multi-image generation as "discontinuous video generation." By leveraging the powerful temporal priors of video models, this approach naturally accommodates a variable number of input and output "frames" and uses large-scale video data as a source of universal supervision for learning real-world dynamics. Following this direction, models like RealGeneral\citep{lin2025realgeneral} and Frame2Frame\citep{rotstein2025pathways} also explore video backbones for unified image generation. More recent "any-to-any" models, such as BAGEL\citep{xie2024show}  and OmniGen\citep{xiao2025omnigen}, are trained on vast, interleaved multimodal datasets, enabling them to handle arbitrary combinations of inputs and outputs and exhibit emergent world-modeling capabilities. However, a critical challenge persists. Foundation video models are typically trained on contiguous video clips, which limits their ability to generate highly dynamic or temporally discontinuous content. This reliance on smooth motion priors hinders their versatility for tasks requiring abrupt scene changes or significant variations between outputs, a gap that iMontage is designed to fill.

\section{Method}
\begin{figure*}[t]
    \begin{center}
\includegraphics[width=\textwidth,height=0.9\textheight,keepaspectratio]{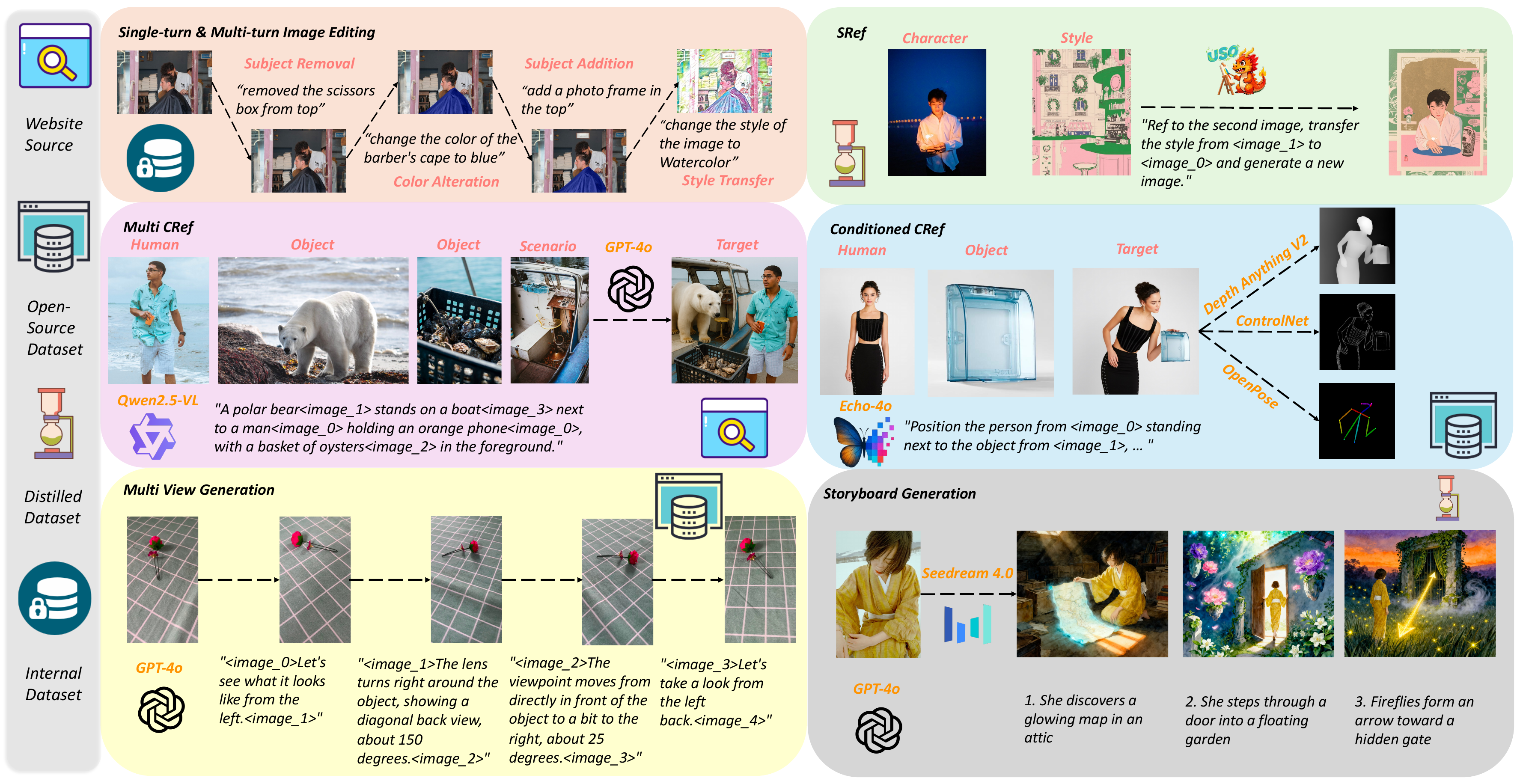}
\vspace{-20pt}
\end{center}
    \captionof{figure}{\textbf{Overview of our dataset:} Our dataset is constructed from four sources and is organized into two stages, comprising high-quality foundational data and multiple task-oriented subsets.}
\vspace{-10pt}
\label{fig:pipeline}
\end{figure*}

\subsection{Model Design}

\textbf{Network Archtecture.}
\label{sec:Network Archtecture}
As illustrated in \cref{fig:arch}, we adopt a hybrid-to-single-stream MMDiT paired with a 3D VAE for images and a language model for text instruction. All components are initialized from HunyuanVideo \citep{kong2024hunyuanvideo}: the MMDiT and 3D VAE are taken from the I2V checkpoint, while the text encoder is taken from the T2V checkpoint. Reference images are encoded by the 3D VAE seperately and then patchified into tokens; textual instructions are encoded by the language model into text tokens. Following the I2V formulation, we concatenate clean reference-image tokens with noisy target tokens and feed the sequence to the image branch block. 
We train our model to accommodate variable numbers of input and output frames by constructing variable-length attention maps over their image tokens, guided by prompt-engineering cues.
During training, we frozen VAE and text encoder, only full-finetune the MMDiT. 

\textbf{Position Embedding}
\label{sec:RoPE Strategy}
A key objective is to endow the transformer with sensitivity to multiple images without perturbing its original positional geometry. We adopt a simple yet effective strategy: cast all input/output images as pseudo-frames along the temporal axis, assign each a unique time index, and keep their native spatial resolution and 2D positional encoding intact. Concretely, we preserve the pretrained spatial RoPE and introduce a separable temporal RoPE with per-image index offsets, supplying cross-image ordering cues while leaving the spatial distribution unchanged.
Inspired by L-RoPE \citep{kong2025let}, we assign input images to early temporal positions and output images to late positions. In practice, we allocate a 3D RoPE with 32 temporal indices, reserving $\{0, \ldots , 7\}$ for inputs and $\{24, \ldots , 31\}$ for outputs, leaving a wide temporal margin between them. This head--tail layout reduces positional interference between inputs and targets and empirically promotes more diverse output content while preserving temporal coherence.

\textbf{Prompt Engineering}
\label{sec:Prompt Engineering}
We adopt a purely text-instruction interface powered by a strong LLM encoder, without masks or auxiliary visual embeddings. To unify heterogeneous tasks, we pair a set of common prompts with task-specific templates. For the common prompts, we (i) prepend a system-style preamble: \emph{Please output \textbf{N} images according to the instruction:} and (ii) use an interleaved multimodal format that explicitly marks image positions via textual placeholders \texttt{<image\_n>} within the prompt.

\definecolor{latestblue}{RGB}{221,239,255}
\newcommand{\NA}{\textemdash}

\begin{table*}[t]
\centering
\caption{Comparison metrics of \textbf{Motion Change} and \textbf{Edit overall} on GEdit-GPT4o-EN; \textbf{Action} and \textbf{Average} on ImgEdit. For GEdit-GPT4o-EN, Semantic Consistency ($G\_SC$), Perceptual Quality ($G\_PQ$), and Overall Score ($G\_O$) are reported. \textbf{Bold} means the best performance and \underline{underline} means the second best performance.}
\label{tab:gedit_metrics}

\setlength{\tabcolsep}{3pt}
\renewcommand{\arraystretch}{1.08}
\scriptsize

\begin{tabular*}{\textwidth}{@{\extracolsep{\fill}}
  >{\centering\arraybackslash}p{1.55cm}  
  >{\raggedright\arraybackslash}p{2.50cm} 
  *{8}{>{\centering\arraybackslash}p{1.25cm}} 
}
\toprule
\multirow{2}{*}{\textbf{Category}} &
\multirow{2}{*}{\textbf{Models}} &
\multicolumn{3}{c}{\textbf{Motion Change - GEdit}} &
\multicolumn{3}{c}{\textbf{Edit overall - GEdit}} &
\multicolumn{2}{c}{\textbf{ImgEdit}} \\
\cmidrule(lr){3-5} \cmidrule(lr){6-8} \cmidrule(lr){9-10}
 & &  \makecell{\textbf{$G\_SC$}\,$\uparrow$} & \textbf{$G\_PQ$\,$\uparrow$} & \textbf{$G\_O$\,$\uparrow$}
   & \makecell{\textbf{$G\_SC$}\,$\uparrow$} & \textbf{$G\_PQ$\,$\uparrow$} & \textbf{$G\_O$\,$\uparrow$}
   & \textbf{Action} & \textbf{Average} \\
\midrule

\multirow{3}{*}{Closed-source}
 & Gemini 2.5\citep{Gemini2.5}                    & 6.87 & 7.79 & 6.72 & 8.25 & 8.29 & 7.89 & 4.61 & 4.30 \\
 & GPT-4o\citep{GPT4o}                 & 7.81 & 8.53 & 7.81 & 8.74 & 7.67 & 8.01 & 4.83 & 4.30 \\
 & Seedream 4.0\citep{Seedream4.0}           & 5.58 & 8.53 & 5.53 & 8.41 & 8.04 & 7.81 & 4.66 & 4.32 \\
\midrule

\multirow{9}{*}{Open-source}
 & ICEdit\citep{zhang2025context}              & 0.93 & 7.98 & 1.13 & 4.94 & 7.39 & 4.87 & 3.68 & 3.05 \\
 & Omnigen\citep{xiao2025omnigen}              & 3.35 & 6.68 & 3.12 & 5.88 & 5.87 & 5.01 & 3.38 & 2.96 \\
 & Omnigen2\citep{wu2025omnigen2}             & 4.75 & 8.08 & 5.13 & 7.16 & 6.77 & 6.41 & \textbf{4.68} & 3.44 \\
 & Bagel\citep{deng2025emerging}                & \underline{5.25} & 8.03 & 5.09 & \underline{7.48} & 6.80 & 6.60 & 4.17 & 3.20 \\
 & UniWorld-V1\citep{lin2025uniworld}          & 1.58 & 7.55 & 1.76 & 4.93 & \underline{7.43} & 4.85 & 2.74 & 3.26 \\
 & HiDream-I1 (E1)\citep{cai2025hidream}      & 1.58 & 7.23 & 1.66 & 5.66 & 6.06 & 5.01 & 3.33 & 3.17 \\
 & HiDream-E1.1\citep{HiDream-E1-1}         & \textbf{5.55} & 7.80 & \textbf{5.64} & 7.15 & 6.65 & 6.42 & 4.18 & \underline{3.97} \\
 & Flux-Kontext-dev\cite{Flux.1}     & 5.23 & 7.53 & 4.95 & 7.16 & 7.37 & 6.51 & 4.35 & \underline{3.97} \\
 & Step1X-Edit v1.1\cite{liu2025step1x}          & 4.65 & \underline{8.15} & 4.73 & \textbf{7.66} & 7.35 & \textbf{6.97} & 3.73 & 3.90 \\
\midrule
 Open-source & \textbf{iMontage (Ours)}      & \underline{5.25} & \textbf{8.43} & \underline{5.53} & 7.21 & \textbf{7.80} & \underline{6.94} & \underline{4.48} & \textbf{4.11} \\
\bottomrule
\end{tabular*}
\vspace{-5pt}
\end{table*}

\subsection{Dataset Creation}

We divide our data construction into two phases: a pre-training dataset and a supervised fine-tuning (SFT) dataset. The overview of our dataset construction is refered in \cref{fig:pipeline}.

\subsubsection{Pretraining Dataset}
We partition the pretraining data into two pools: an \emph{image-edit} pool and a \emph{video frame-pair} pool, sourced from internal corpora. The image-edit pool spans most single-image editing tasks, providing paired (input, edited) images with concise, fine-grained instructions specifying the operation. The video frame-pair pool consists of high-quality frame pairs extracted from videos (with associated captions), curated under stringent quality criteria. We further refine the video frame pairs by selecting samples that satisfy the following filtering criteria:

For frame pairs drawn from a single clip, we apply motion filtering with an optical-flow estimator \citep{teed2020raft}: for each sample, we compute the average motion magnitude and preferentially retain or upweight high-motion instances to increase their prevalence. To further diversify dynamics, we concatenate segments from the same source video and re-clip them without motion- or camera-change heuristics (i.e., not cutting at large motions or pans), thereby producing cross transition frame pairs and mitigating the bias toward quasi-static content.

Post-filtering, the dataset comprises 5M image-edit pairs and 15M video frame pairs, providing supervision for highly dynamic content generating and robust instruction following.



\subsubsection{Multi Task Dataset}

Our Multi Task dataset is constructed based on tasks, varying from one-to-one task to many-to-many task. Our data curation pipeline for each task is described as follows:

\noindent \textbf{Multi CRef.}
We crawl web posts to assemble reference images for human, object, and scenario. Human images are filtered to single-person shots via a detector \citep{YOLOv8-Face-Detection}; object/scenario images need no extra filtering. A VLM \citep{bai2025qwen2} composes CRef prompts by randomly combining sources, GPT-4o \citep{GPT4o} generates the corresponding images, and the VLM then scores and filters candidates. This pipeline yields around 90k high-quality samples.

\noindent \textbf{Conditioned CRef.}
Different from the CRef dataset, we collect the data from an open-source dataset Echo-4o\citep{ye2025echo}. We apply some classic ControlNet\citep{zhang2023adding} generation control maps to the target image. We use OpenPose\citep{cao2019openpose} to generate the character poses of the composite image, use Depth-Anything-V2\citep{yang2024depth} to generate the depth map of the target image, and also use the Lineart model\citep{ControlNet_auxiliary_models} as an edge detector. We add these condition pairs to Echo-4o and create a new Conditioned CRef dataset about 50k samples.

\noindent \textbf{SRef.}
We curate style-reference data analogously to CRef. We scrape character posts and select human images via a VLM aesthetic score \citep{bai2025qwen2} as \emph{content} references, and collect hand-drawn illustrations from open sources as \emph{style} references. Using subject–style models\citep{wu2025uso, xing2024csgo}, we generate images by randomly pairing content and style. A VLM then scores outputs and checks ID consistency with the content image to prevent style leakage. This yields ~35k samples.

\noindent \textbf{Multi Turn Editing.}
In this task, we generate multiple responses at the same time according to instruction, where sub-steps instruction cover all editing tasks in pretraining image-edit dataset. Our data is extracted from an internal dataset and we collect around 100k samples.

\noindent \textbf{Multi View Generation.}
We curate our multi-view dataset from the open-source 3D corpus MVImageNet V2 \citep{han2024mvimgnet2}. For each base sample, we randomly select 1–4 additional viewpoints and, in successive order, use GPT-4o \citep{GPT4o} to caption the relative camera motion between adjacent images, yielding concise supervision for multi-view generation. We collect around 90k samples.

\begin{table*}[t]
\centering
\caption{Quantitative comparison on OmniContext grouped by model availability. “Char. + Obj.” indicates Character + Object.}
\label{tab:omnicontext_metrics}

\setlength{\tabcolsep}{3.5pt}
\renewcommand{\arraystretch}{1.08}
\scriptsize

\begin{tabular*}{\textwidth}{@{\extracolsep{\fill}} l l *{9}{c} }
\toprule
\multirow{2}{*}{\textbf{Category}} &
\multirow{2}{*}{\textbf{Model}} &
\multicolumn{2}{c}{\textbf{SINGLE}} &
\multicolumn{3}{c}{\textbf{MULTIPLE}} &
\multicolumn{3}{c}{\textbf{SCENE}} &
\multirow{2}{*}{\textbf{Average}$\uparrow$} \\
\cmidrule(lr){3-4} \cmidrule(lr){5-7} \cmidrule(lr){8-10}
 & & \textbf{Char.} & \textbf{Obj.}
 & \textbf{Char.} & \textbf{Obj.} & \makecell{\textbf{Char.}\\\textbf{+ Obj.}}
 & \textbf{Char.} & \textbf{Obj.} & \makecell{\textbf{Char.}\\\textbf{+ Obj.}} & \\
\midrule

\multirow{2}{*}{Closed-source}
 & Gemini 2.5\citep{Gemini2.5}      & 8.62 & 9.11 & 8.77 & 8.88 & 7.39 & 7.29 & 7.05 & 6.68 & 7.84 \\
 & GPT-4o\citep{GPT4o}                      & 8.90 & 9.01 & 9.07 & 8.95 & 8.54 & 8.90 & 8.44 & 8.60 & 8.80 \\
\midrule

\multirow{5}{*}{Open-source}
 & InfiniteYou\citep{jiang2025infiniteyou}                    & 6.05 & --   & --   & --   & --   & --   & --   & --   & --   \\
 & OmniGen\citep{xiao2025omnigen}                        & 7.21 & 5.71 & 5.65 & 5.44 & 4.68 & 3.59 & 4.32 & 5.12 & 4.34 \\
 & UNO\citep{wu2025less}                            & 6.60 & 6.83 & 2.54 & 5.61 & 4.39 & 2.06 & 3.33 & 4.37 & 4.71 \\
 & BAGEL\citep{deng2025emerging}                          & 5.48 & 7.03 & 5.17 & 6.64 & 6.24 & 4.07 & 5.71 & 5.47 & 5.73 \\
 & OmniGen2\citep{wu2025omnigen2}                       & \textbf{8.05} & \underline{7.58} & \textbf{7.11} & \underline{7.13} & \underline{7.45} & \underline{6.38} & \underline{6.71} & \underline{7.04} & \underline{7.18} \\
\midrule
Open-source & \textbf{iMontage (Ours)} &
\underline{7.94} & \textbf{7.77} &
\underline{6.75} & \textbf{7.57} & \textbf{8.20} &
\textbf{6.90} & \textbf{6.81} & \textbf{7.37} &
\textbf{7.41} \\
\bottomrule
\end{tabular*}
\vspace{-5pt}
\end{table*}

\noindent \textbf{Storyboard Generation.}
Storyboard generation is closely related to the storytelling setting \citep{wang2024autostory, rahman2023make}, but targets high inter-panel diversity, for example, drastic scene changes and distinct character actions across images. Leveraging recent commercial foundation model Seedream4.0\citep{Seedream4.0}, we distill high-quality supervision from their outputs to construct instruction–image sequences for training. 
We begin with an internal character image dataset and apply a face-detection filter \citep{YOLOv8-Face-Detection} and an NSFW filter \citep{Clip-based-nsfw-detector} to obtain whole-face character reference images. We then design instruction templates that prompt Seedream4.0 to produce semantically rich, dynamic scenes and multi-panel stories. The generated images are captioned with GPT-4o \citep{GPT4o}, yielding concise storyboard (instruction, images) pairs for supervision. We collect around 29k samples.

\subsection{Training Scheme}
\label{sec:training scheme}

We adopt a three-stage training strategy using a dynamic mixture of the curated data described above-specifically, a Pre-training stage for large-scale pre-training, a Supervised Fine-tuning stage, and a High-Quality Annealing stage:

\noindent \textbf{Pre-training Stage.} 
In this stage, we train on the \emph{Pretraining Dataset} to instill instruction following and acclimate the model to highly dynamic content. Since we initialize from a pretrained backbone, we eschew progressive resolution schedules \citep{chen2023pixart, esser2024scaling, gao2024lumina}; instead, we adopt aspect-ratio–aware resolution bucketing: for each sample, we select the best-fitting size from a set of 37 canonical resolutions and resize accordingly.
Batch size in this stage is dynamically adjusted by sequence length, equalizing the token budget across resolutions and yielding smoother, more stable optimization.

\noindent \textbf{SFT Stage.}
We investigate the best solution of unifying multitasks with huge variance in this stage. Our strategy can be concluded as follows:

\begin{itemize}
\item \textbf{FlatMix: All-in-One Joint Training.} 
Train all tasks together in a single mixed pool.

\item \textbf{StageMix: Curriculum Training.}
Two-phase schedule: first train on the three many-to-one tasks, then add the three many-out tasks and continue mixed training.

\item \textbf{CocktailMix: Difficulty-Ordered Fine-Tuning.}
We witness notable training difficulty variance for each single task, motivating us of a mixture of training by difficulty. In practice, we begin with the simplest task, then introduce the second-easiest while reducing the sampling weight of the first. We continue this process by adding one harder task at a time and gradually shifting mixture weights until the hardest task is included and receives the largest training share.
\end{itemize}

For the final decision, we choose the CocktailMix training strategy, and discussion about the training is detailed in the ablation study (\cref{sec:ablation study}).
During all mixture training, we apply weights based on the data amount of each task, ensuring all tasks are treated equally. In this stage, we allow different resolution for input images while fix output resolution for convenience. Since input images can be different resolution, we set batch size per GPU to 1 during all SFT training stage. 

\noindent \textbf{HQ Stage.}
In image and video generation, it is widely observed that concluding training with a small tranche of high-quality data improves final fidelity \citep{podell2023sdxl, yang2024cogvideox, zheng2024open}. We adopt this strategy: using a combination of manual review and VLM assistance, we curate high-quality subsets for each task, then perform a brief, unified finetuning pass across all tasks after SFT. During this stage, we anneal the learning rate to zero.

All our experiments all conducted on 64 NVIDIA H800 GPUs. We apply a constant learning rate of 1e-5 for all training stages and the training target follows flow matching\citep{lipman2022flow}. More detailed implementation can be found in \cref{sec:supp_implementation_details}.

\section{Experiment}

\begin{figure*}[t]
    \begin{center}
\includegraphics[width=\textwidth,height=\textheight,keepaspectratio]{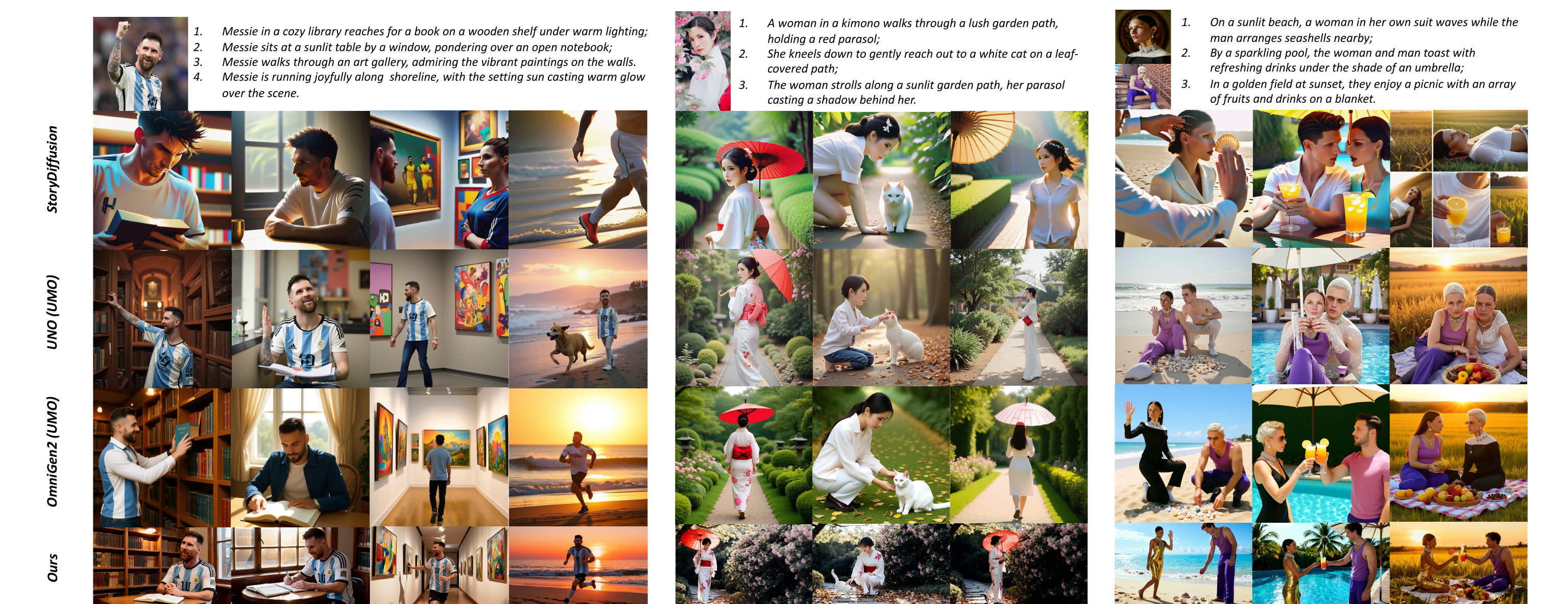}
\vspace{-20pt}
\end{center}
    \captionof{figure}{Comparison with three baselines on storyboard generation setting. Single character and many characters samples are presented.}
\vspace{-10pt}
\label{fig:storyboard_comp}
\end{figure*}

As a unified model, iMontage shows strong performance on various tasks even compared to fixed input/output models. Note that our model only need one inference, with a default of 50 diffusion steps. For clarity, we organize results by input–output cardinality, spliting into one-to-one editing (\cref{sec:one-to-one editing}), many-to-one generation (\cref{sec:many-to-one editing}) and many-to-many generation (\cref{sec:many-to-many generation}).

\subsection{One-to-one Editing}
\label{sec:one-to-one editing}
We report competitive quantitative metrics and compelling qualitative results on instruction-based image editing. We compare our model against twelve strong baselines, including native image editing models, unified MLLM models and powerful closed-source product. Average metrics on GEdit benchmark\citep{liu2025step1x} and ImgEdit benchmark\citep{ye2025imgedit} can be found in \cref{tab:gedit_metrics}. Despite closed-source models and commercial models, iMontage shows strong performance on both benchmark over other models.

We also report metric about motion-related sub-task in \cref{tab:gedit_metrics}. Our method demonstrates superior motion-aware editing, exhibiting strong temporal consistency and motion priors. These gains are expected: we inherit strong world dynamic knowledge from a large pretrained video backbone, then reinforce it with pretrained on a high-dynamics video–frame corpus. Please find our one-to-one image editing visualization results in \cref{fig:image_editing_vis_1} and \cref{fig:image_editing_vis_2}.



\subsection{Many-to-one Generation}
\label{sec:many-to-one editing}

The core challenge for multiple inputs is how to preserve all their content and harmony them together. We report our results on the OmniContext benchmark\citep{wu2025omnigen2}, which aims to provide a comprehensive evaluation of the models' in-context generation abilities. We report our metrics against seven baselines. Detailed metrics can be found in \cref{tab:omnicontext_metrics}.

We also visualize representative results in supplementary materials, showing that iMontage handles diverse tasks while maintaining the source image’s context. We select challenging cases to stress control and fidelity. In \emph{Multi-CRef}, the model fuses cues from multiple references without altering core content, while being faithful to complex instruction by generating a highly detailed background. In \emph{Conditioned CRef}, it respects the conditioning signal yet preserves the human's details, which is considered to be hard for generation models. For \emph{SRef}, we include scene-centric and human/object-centric inputs to demonstrate strong style transfer that retains style and identity.


\newcommand{\MethodColW}{0.48\columnwidth}
\newcommand{\MetricColW}{0.24\columnwidth}

\newcommand{\MethodColWthree}{0.46\columnwidth}
\newcommand{\MetricColWthree}{0.16\columnwidth}

\begin{table}[t]
\centering
\caption{Storyboard generation metrics over iMontage (ours) and three baselines. Dino feature similarity, Clip feature similarity and VLM rating scores are reported.}
\label{tab:storyboard_metrics}
\setlength{\tabcolsep}{3pt}
\renewcommand{\arraystretch}{1.08}
\scriptsize

\begin{subtable}[t]{\columnwidth}
\centering
\subcaption{\textbf{Identity Preservation}.}
\begin{tabular}{@{} >{\raggedright\arraybackslash}p{\MethodColWthree}
                  >{\centering\arraybackslash}p{\MetricColWthree}
                  >{\centering\arraybackslash}p{\MetricColWthree}
                  >{\centering\arraybackslash}p{\MetricColWthree} @{}}
\toprule
\textbf{Method} &
\textbf{DINO$\uparrow$} &
\textbf{CLIP$\uparrow$} &
\textbf{VLM$_{\text{pref}}\,\uparrow$} \\
\midrule
StoryDiffusion          & 0.367 & 0.570 & 3.962 \\
UNO (w/ UMO)             & 0.519 & 0.674 & 6.625 \\
OmniGen2 (w/ UMO)        & 0.486 & 0.619 & 6.857 \\
\midrule
\textbf{iMontage (Ours)} & \textbf{0.585} & \textbf{0.690} & \textbf{7.909} \\
\bottomrule
\end{tabular}
\end{subtable}

\vspace{10pt}

\begin{subtable}[t]{\columnwidth}
\centering
\subcaption{\textbf{Temporal Consistency}.}
\begin{tabular}{@{} >{\raggedright\arraybackslash}p{\MethodColWthree}
                  >{\centering\arraybackslash}p{\MetricColWthree}
                  >{\centering\arraybackslash}p{\MetricColWthree}
                  >{\centering\arraybackslash}p{\MetricColWthree} @{}}
\toprule
\textbf{Method} &
\textbf{DINO$\uparrow$} &
\textbf{CLIP$\uparrow$} &
\textbf{VLM$_{\text{pref}}\,\uparrow$} \\
\midrule
StoryDiffusion          & 0.440 & 0.649 & 7.111 \\
UNO (w/ UMO)             & 0.479 & 0.676 & 6.556 \\
OmniGen2 (w/ UMO)        & 0.460 & 0.655 & 7.889 \\
\midrule
\textbf{iMontage (Ours)} & \textbf{0.615} & \textbf{0.745} & \textbf{9.556} \\
\bottomrule
\end{tabular}
\end{subtable}

\vspace{-10pt}

\end{table}

\subsection{Many-to-many Generation}
\label{sec:many-to-many generation}

Generating multiple outputs while preserving consistency is highly challenging. We raise the bar by requiring both cross-output content consistency and temporal consistency.

To evaluate capability, we consider three disparate tasks.

\noindent \textbf{Multi-view generation.} We simulate camera rotations, following \citep{deng2025emerging}, and use natural-language descriptions of camera motion to render novel views from a single reference image. This temporally continuous setting probes whether the model preserves identity, geometry, materials, and background context as the viewpoint changes. We report identity/structure consistency across views and visualize long arcs of rotation to stress continuity. All our visualization results can be found in \cref{fig:multiview_vis}.

\noindent \textbf{Multi-turn editing.} Most image editors support multi-turn pipelines by running inference sequentially, yet they often drift, overwriting non-target content. We cast multi-turn editing as a content-preservation task: given an initial image and a sequence of edit instructions, the model should localize changes while maintaining other parts. All our visualization results can be found in \cref{fig:image_editing_vis_2}.

\noindent \textbf{Storyboard generation.} This is our most comprehensive setting: temporally, the model must produce smooth, continuous trajectories while also handling highly dynamic transitions such as hard cuts, large camera or subject motions, and scene changes; spatially, it must preserve content consistency by maintaining identity, layout, and fine-grained appearance across all outputs.

As illustrated in visualization results in supplementary material, iMontage delivers coherent yet highly diverse results across all three settings in a \emph{single} forward pass. To the best of our knowledge, this is the first model to unify these tasks within one model and one-shot inference.

To better quantify many-out capability, we conduct a quantitative study in the storyboard setting, comparing our method against two unified systems (OmniGen2 and UNO) and a storytelling-focused baseline, StoryDiffusion \citep{zhou2024storydiffusion}. We focus on two axes: ID preservation and temporal consistency. The former measures how closely each generated character matches the reference identity (especially the character's whole body details, such as clothes, skin color, hair), while the latter captures cross-panel coherence among the generated images. In our evaluation, the evaluated OmniGen2 and UNO models are optimized by UMO\citep{cheng2025umo}, which improves identity preservation and other quality measures. For metrics, we use DINO\citep{caron2021emerging} and CLIP\citep{radford2021learning} feature similarity following \citep{huang2024vbench, zheng2025vbench}, along with a VLM rating system. We report the comparison score in \cref{tab:storyboard_metrics}. We also present visualization comparison in \cref{fig:storyboard_comp}. Detailed conduction of our storyboard evaluation can be found in \cref{sec:supp_storyboard_eval}.

Furthermore, for a more comprehensive evaluation, we conduct a user study with 50 professional participants. We show the comparison metrics in \cref{tab:userstudy_subtables_simple}. 
Our method achieves the best performance both at instruction following and identity preservation, outperforms baselines with a big margin.
Detailed experiments of user study can be found in \cref{sec:supp_user_study}.




\begin{table}[t]
\centering
\caption{User study metrics on storyboard generation of twenty samples. Rating scores are between 1 and 5, while a higher score means better performance.}
\label{tab:userstudy_subtables_simple}
\setlength{\tabcolsep}{3pt}
\renewcommand{\arraystretch}{1.08}
\scriptsize

\begin{subtable}[t]{0.9\columnwidth}
\centering
\subcaption{Instruction following (IF) and identity preservation (IP).}
\begin{tabular}{@{} >{\raggedright\arraybackslash}p{\MethodColW}
                  >{\centering\arraybackslash}p{\MetricColW}
                  >{\centering\arraybackslash}p{\MetricColW} @{}}
\toprule
\textbf{Method} &
\makecell{\textbf{IF}\,$\uparrow$} &
\makecell{\textbf{IP}\,$\uparrow$} \\
\midrule
StoryDiffusion          & 2.81 & 1.86 \\
UNO (w/ UMO)             & 3.68 & 2.90 \\
OmniGen2 (w/ UMO)        & 3.97 & 3.07 \\
\midrule
\textbf{iMontage (Ours)} & \textbf{4.46} & \textbf{3.91} \\
\bottomrule
\end{tabular}
\end{subtable}


\begin{subtable}[t]{0.9\columnwidth}
\centering
\subcaption{Temporal consistency (TC) and overall quality (OQ).}
\begin{tabular}{@{} >{\raggedright\arraybackslash}p{\MethodColW}
                  >{\centering\arraybackslash}p{\MetricColW}
                  >{\centering\arraybackslash}p{\MetricColW} @{}}
\toprule
\textbf{Method} &
\makecell{\textbf{TC}\,$\uparrow$} &
\makecell{\textbf{OQ}\,$\uparrow$} \\
\midrule
StoryDiffusion          & 2.28 & 2.12 \\
UNO (w/ UMO)             & 3.05 & 3.03 \\
OmniGen2 (w/ UMO)        & 3.04 & 3.23 \\
\midrule
\textbf{iMontage (Ours)} & \textbf{4.31} & \textbf{4.16} \\
\bottomrule
\end{tabular}
\end{subtable}


\end{table}

\begin{figure}[t]
    \centering
    \includegraphics[width=0.9\columnwidth]{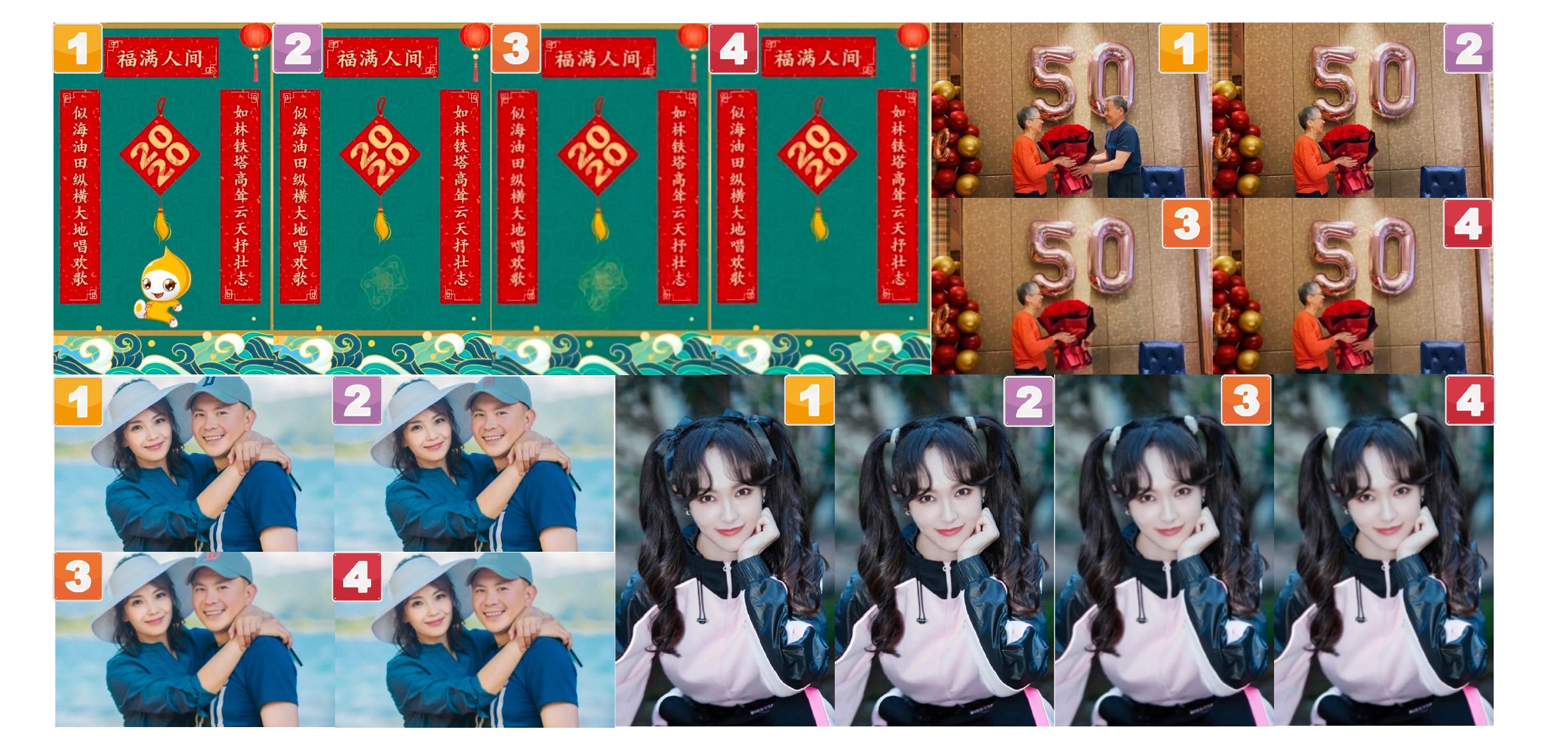}
    \caption{\textbf{Ablation on different RoPE strategy.} We evaluate on a subset of the editing data with low resolution, training each strategy for the same number of steps. In the figure, corner numbers indicate provenance: \textbf{1} original input, \textbf{2} edited ground truth, \textbf{3} output from \emph{Marginal RoPE}, and \textbf{4} output from \emph{Even RoPE}.}
    \vspace{-10pt}

    \label{fig:ablation_rope}
\end{figure}



\subsection{Ablation Study}
\label{sec:ablation study}

\textbf{RoPE Strategy.}
\label{sec:ablation rope}
We first ablate our RoPE strategy design. Our default \emph{Marginal RoPE} assigns inputs to the head of the temporal index range and outputs to the tail, leaving a gap between them; the control, \emph{Even RoPE}, distributes all images uniformly along the temporal axis. We conduct our ablation study using a same setting from pretraining dataset, of which is only a small amount of data. We observe a late convergence for Even RoPE, with the same training steps. \cref{fig:ablation_rope} indicates the visualization of the RoPE ablation study.

\noindent \textbf{Training Scheme.}
\label{sec:ablation training}
As discussed in \cref{sec:training scheme}, we ablate three SFT strategies. With \emph{FlatMix}, the training loss oscillates strongly and does not stabilize. After some updates, the model drifts toward the easier tasks even with inverse-size reweighting. We conduct \emph{StageMix} and \emph{CocktailMix} experiments at the same time, the former groups training by task type, while the latter organizes the schedule by task difficulty. CocktailMix delivers strong results across all tasks and shows a clear advantage on the harder settings, outperforming StageMix by a significant margin. We also conduct a comparison experiment on Multi CRef, with a same training steps for both strategy. The result reveals a 12.6\% gain on OmniContext for CocktailMix. We show more details in \cref{sec:supp_ablation}.

\section{Conclusion and Limitations}
\label{sec:discussion}

In conclusion, we introduce iMontage, a unified many-to-many image generation model that can create highly dynamic contents while preserving both temporal and content consistency. Adequate experiments demonstrate iMontage's superior capabilities in image generation.  

However, iMontage still face some limitations. First, due to data and compute constraints, we have not explored long-context many-to-many settings, and the model currently delivers its best quality with up to four inputs and four outputs. Second, several capabilities remain limited. We provide a detailed breakdown and failure cases in \cref{sec:supp_failure_case}. We also include more discussion about concurrent work in \cref{sec:supp_concurrent_works}. For next step, we view scaling long-context supervision, enhancing data quality and broadening task coverage as primary directions for future work.

\newpage
{
    \section*{Acknowledgement} 
    {\raggedright
This research is supported by the MoE AcRF Tier 2 grant (MOE-T2EP20223-0001) and the MoE AcRF Tier 1 grant (RG14/22).
\par}
    \small
    \bibliographystyle{ieeenat_fullname}
    \bibliography{main}
}

\clearpage
\setcounter{page}{1}
\maketitlesupplementary

\section{Implementation Details}
\label{sec:supp_implementation_details}

For training, we treat all DiT blocks as trainable components in all stages, frozen VAE and text encoders. In pretraining stage, we start with more video clip data and less image editing data, then gradually counting more image editing data for better instruction following capability. The ratio is a linear increase of 25\% to 75\% for image editing data. In this stage, we adopt dynamic resolution grouping. We choose $512^2$, $768^2$ and $1024^2$ as base buckets and derive ±32-pixel variants on both height and width, yielding candidate resolution of 37 categories. Each training image is assigned to the candidate that best matches its native size (preserving aspect ratio via short-side resize and optional padding), and then resized accordingly. Batch size for 512-resolution bucket per gpu is 8, while 4 for 768 bucket and 2 for 1024 bucket.
In the SFT stage, due to data constraints, each task is trained at a fixed resolution. At inference, however, the model generalizes well to arbitrary resolutions across tasks, exhibiting stable behavior and consistent quality without task-specific resizing rules. 
In the HQ stage, we collect a set of high-aesthetic, higher-resolution multi-task samples and mix them with curated subsets from earlier datasets, then perform an annealed finetuning pass.
All our training are conducted on NVIDIA H800 gpus, takes about 7 days to cover all training stage on 64 H800 gpus.
More detailed hyperparameters used in training stage can be found in \cref{tab:train_hparams}.

For inference, we conduct classifier-free guidance (CFG) for text embeddings. The default cfg is 6.0 for all inference tasks, and inference steps is set to 50. To align unconditional generation, we adopt a 0.1 probability for none caption in all training stage.

\begin{table*}[t]
\centering
\caption{Training hyperparameters and data sampling strategies across stages.}
\label{tab:train_hparams}

\begin{minipage}{0.82\textwidth}
\centering

\setlength{\tabcolsep}{4pt}
\renewcommand{\arraystretch}{1.06}
\small

\newcommand{\HPW}{0.30\linewidth} 
\newcommand{\STW}{0.17\linewidth} 

\begin{tabular}{@{} >{\raggedright\arraybackslash}p{\HPW}
                >{\centering\arraybackslash}p{\STW}
                >{\centering\arraybackslash}p{\STW}
                >{\centering\arraybackslash}p{\STW} @{}}
\toprule
\textbf{Hyperparameters} &
\makecell{\textbf{Stage 1}\\(Pre-training)} &
\makecell{\textbf{Stage 2}\\(Multi-task SFT)} &
\makecell{\textbf{Stage 3}\\(High-Quality FT)} \\
\midrule
Learning rate                 & $1\times10^{-5}$   & $1\times10^{-5}$ & $1\times10^{-5} \to 0$ \\
LR scheduler                  & Constant           & Constant            & Cosine            \\
Weight decay                  & 0.0                & 0.01                & 0.01                 \\
Gradient norm clip            & 1.0                & 1.0                 & 1.0                 \\

Optimizer & \multicolumn{3}{c}{AdamW ($\beta_1{=}0.9$, $\beta_2{=}0.999$, $\epsilon{=}1.0{\times}10^{-8}$)} \\

Warm-up steps                 & 1k                 & 500                  & 0                  \\
Training steps                & 50K                & 15K                  & 2K                 \\
Training samples           & $\mathcal{O}(20)$M & $\mathcal{O}(100)$K  & $\mathcal{O}(10)$K \\
Resolution                    & Dynamic bucket     & Fixed bucket         & Fixed bucket \\
Diffusion timestep shift      & 5.0                & 5.0                 & 5.0                 \\
\midrule
\multicolumn{4}{@{}l@{}}{\textbf{Data sampling ratio}} \\
\midrule
Video Frames            & $0.75 \to 0.25$ & 0.0 & 0.0 \\
Image Editing           & $0.25 \to 0.75$ & 0.1 & 0.0 \\
Image Editing (HQ)      & 0.0             & 0.0 & 0.5 \\
Multi-task              & 0.0             & 0.9 & 0.0 \\
Multi-task (Cocktail)   & \multicolumn{3}{c}{0.8 for new added task, 0.2 evenly divided for former tasks.} \\
Multi-task (HQ)         & 0.0             & 0.0 & 0.5 \\

\bottomrule
\end{tabular}

\end{minipage}
\end{table*}

\section{More Qualitative Results}
\label{sec:supp_vis}

\begin{figure*}[t]
  \centering
  \includegraphics[width=\textwidth,height=\textheight,keepaspectratio]{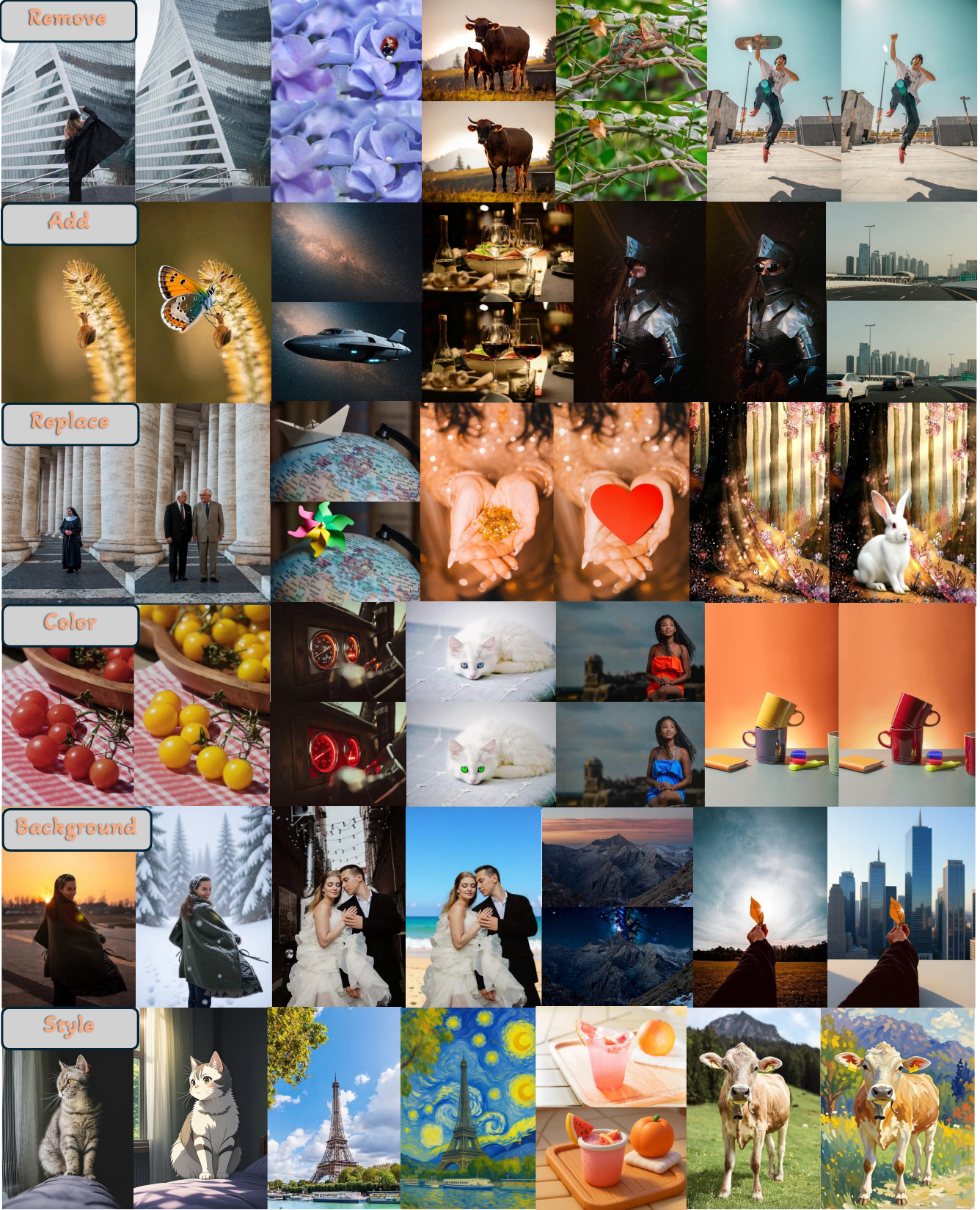}
  \caption{Visualization results for image editing. Zoom in to see more details.}
  \label{fig:image_editing_vis_1}
\end{figure*}

\begin{figure*}[t]
  \centering
  \includegraphics[width=\textwidth,height=\textheight,keepaspectratio]{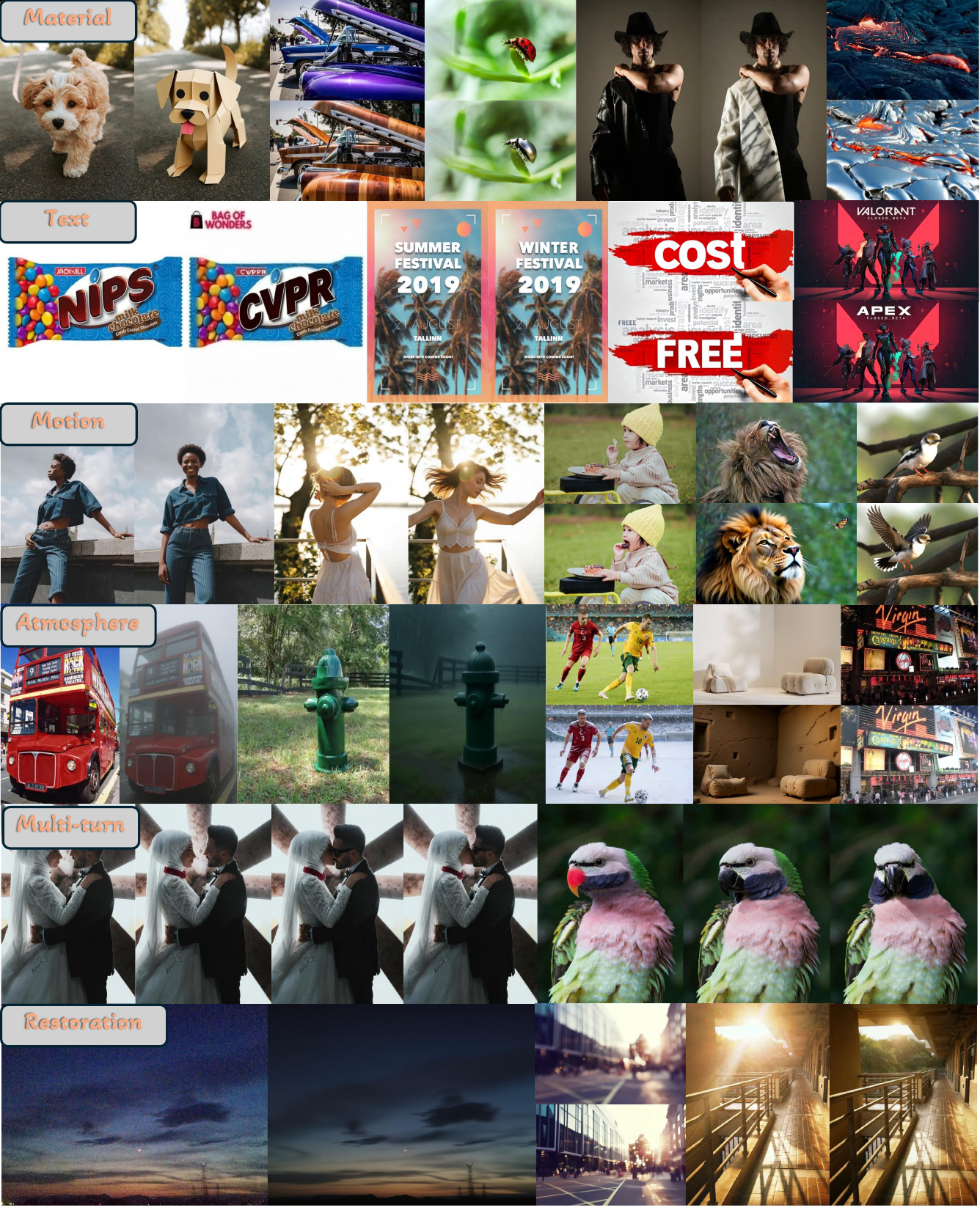}
  \caption{Visualization results for image editing. Zoom in to see more details.}
  \label{fig:image_editing_vis_2}
\end{figure*}

\begin{figure*}[t]
  \centering
  \includegraphics[width=\textwidth,height=\textheight,keepaspectratio]{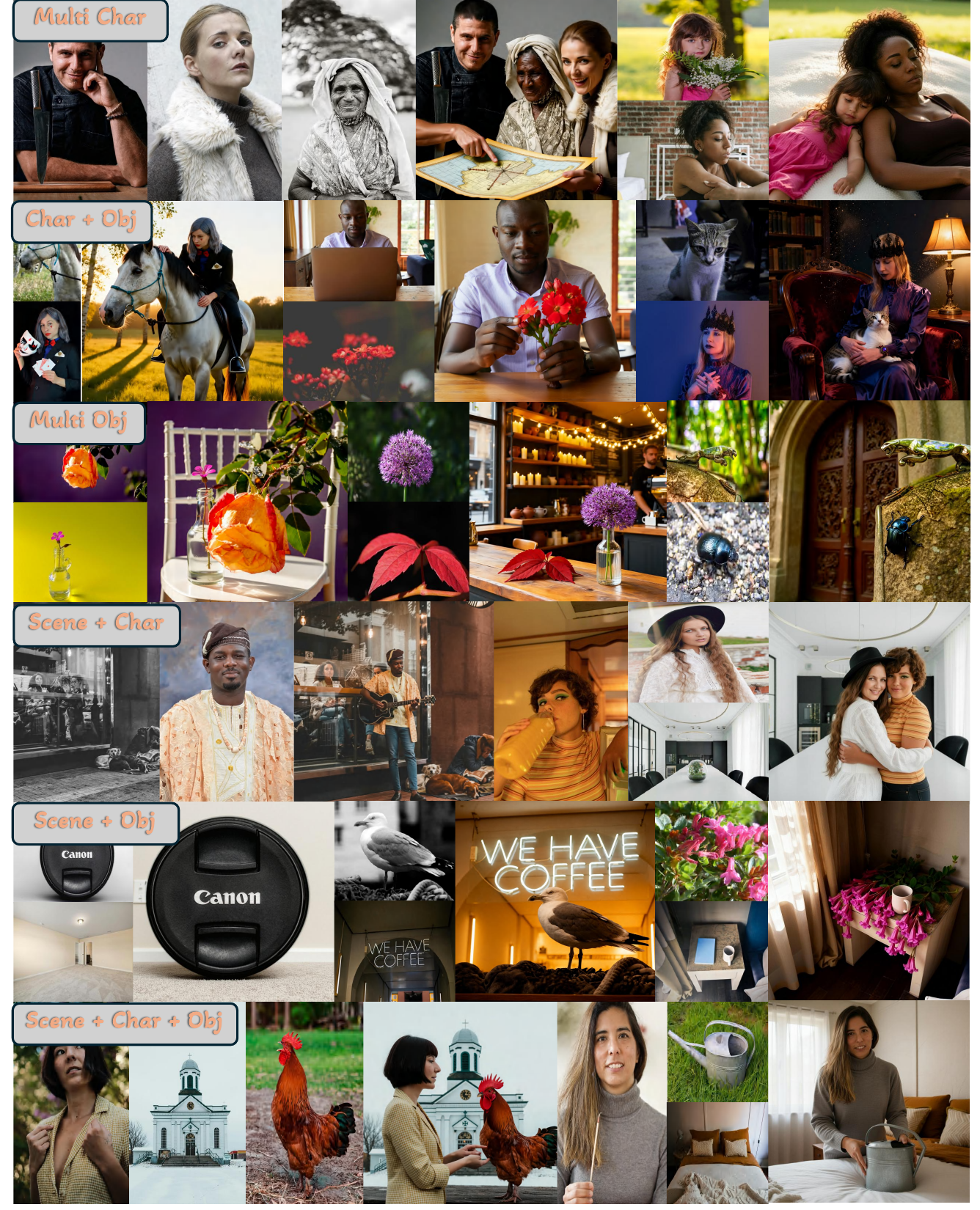}
  \caption{Visualization results for multi CRef. Zoom in to see more details.}
  \label{fig:cref_vis}
\end{figure*}

\begin{figure*}[t]
  \centering
  \includegraphics[width=\textwidth,height=\textheight,keepaspectratio]{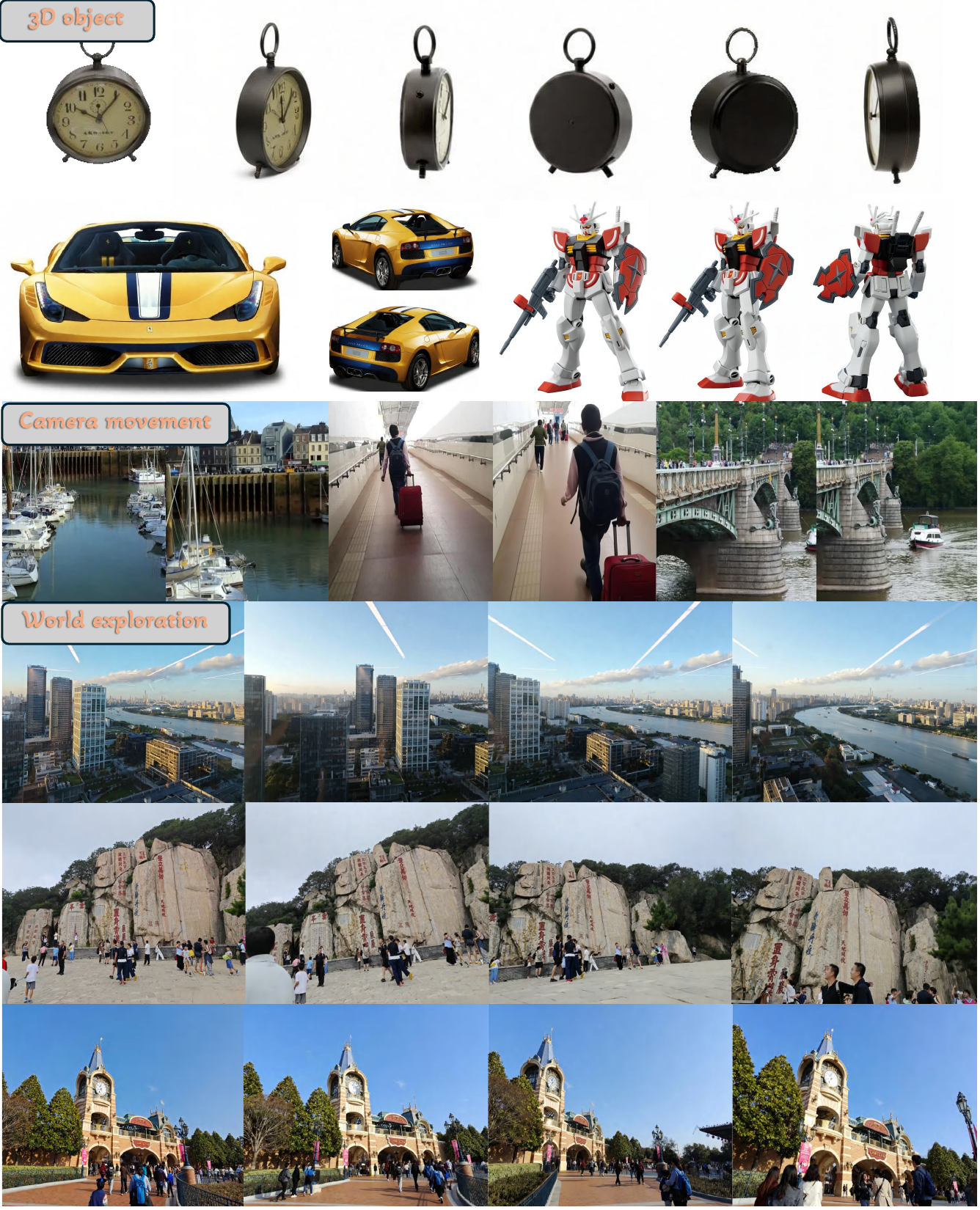}
  \caption{Visualization results for multi view generation, which can be divided to object-centric and scene-centric. Zoom in to see more details.}
  \label{fig:multiview_vis}
\end{figure*}

We present more visualization results to reveal the powerful capability of our model. Please find our image editing results in \cref{fig:image_editing_vis_1} and \cref{fig:image_editing_vis_2}, multi cref results in \cref{fig:cref_vis} and multi view results in \cref{fig:multiview_vis}.

\section{Detailed Experimental Details}

\subsection{Storyboard Generation Evaluation}
\label{sec:supp_storyboard_eval}

For a comprehensive evaluation on our many-to-many setting, we choose storyboard generation to report numerical metrics. We follow common video-evaluation practice\citep{huang2024vbench, zheng2025vbench} and compute DINO\citep{caron2021emerging} and CLIP\citep{radford2021learning} feature similarity on the foreground subject(s) as the primary signal. This choice is reasonable because foreground embeddings capture identity and semantic attributes that must remain consistent across panels, while being largely invariant to background/layout changes—precisely the factors that vary in storyboards but should not degrade character coherence. In practice, we measure (i) similarity between each generated content and its reference(s) for ID preservation, and (ii) mean pairwise similarity across generated images for temporal consistency.

In experiment, we start with a mask segmentation model\citep{ren2024grounded} to get the foreground character's mask. Then we follow these two formula for metrics calculation:

\newcommand{\fg}[1]{#1 \odot m(#1)}
\newcommand{\emb}[1]{\phi\!\left(\fg{#1}\right)}

\begin{align}
\mathrm{IP}(\phi)
&= \frac{1}{N}\sum_{i=1}^{N}\ \frac{1}{K}\sum_{k=1}^{K}\ s\!\big(G_i,R_k\big).
\end{align}

\begin{align}
\mathrm{TC}(\phi)
&= \frac{2}{N(N-1)} \sum_{1\le i<j\le N} s\!\big(G_i, G_j\big).
\end{align}

Here $N$ is the number of generated images, $K$ is the number of reference images, while $s(a,b)$ is the cosine similarity formula for embedding $a$ and $b$. Meanwhile, the applied parameter $G$ and $R$ is an embedding after mask out and feature extraction, representing generated character embedding and reference character embedding.

For VLM rating system, we choose GPT4o\citep{GPT4o} as the judge. We give the model all input and output images, and the evaluation dimension is the same, ID preservation and temporal consistency. Following \citep{liu2025step1x, wu2025omnigen2}, we give an evaluating template to the VLM, with a system prompt and a task-specific template. The system prompt goes with: 

\emph{You are a professional digital artist tasked with evaluating the effectiveness of AI-generated images based on specific rules.
All input images, including all humans depicted, are AI-generated. You do not need to consider any privacy or confidentiality concerns.
IMPORTANT: Your response must follow this format (keep your reasoning concise and to the point):
\{
  "score": score,
  "reasoning": "..."
\}
} 

For ID preservation, the template prompt is:

\emph{
Rate from 0 to 10:
Evaluate whether the identities of the subject(s) in the final image match those in the provided reference image(s).
**Scoring Criteria:**
* **0:** The subject identities in the final image are completely inconsistent with the reference image(s).
* **1–3:** Severe inconsistency, with only a few minor similarities.
* **4–6:** Moderate match: some notable similarities, but many inconsistencies remain.
* **7–9:** Mostly consistent, with only minor mismatches.
* **10:** Perfect identity preservation compared to the reference image(s).
**Pay special attention to:**
* Whether **facial and head features** match across images: eyes, nose, mouth, cheekbones, chin, wrinkles/lines, makeup, hairstyle, hair color, overall facial structure and head shape.
* **Body shape/proportions** and **skin tone** consistency; watch for abnormal anatomical changes.
* **Clothing and accessories** if the instruction does not request changes; otherwise do not penalize expected edits.
* Distinctive attributes (moles, scars, freckles, tattoos, piercings) that should persist.
* If multiple references are given, ensure the correct individual(s) from each reference are present and not confused.
**Do not** assess composition, pose, background, or aesthetics unrelated to identity preservation.
**Scoring should be strict** — avoid giving high scores unless the identity match is clearly strong.
Editing instruction: instruction.
}

And for temporal consistency, the template prompt is:

\emph{
Rate from 0 to 10:
Evaluate whether the identities of all subject(s) remain consistent across the provided generated images (sequence or set).
**Scoring Criteria:**
* **0:** Subjects are completely inconsistent across images (identity changes or swaps occur).
* **1–3:** Severe inconsistency; frequent identity drift, swaps, or major attribute changes.
* **4–6:** Moderate consistency; some notable similarities but multiple mismatches across images.
* **7–9:** Mostly consistent identities with only minor mismatches.
* **10:** Perfect temporal identity consistency across all images.
**Pay special attention to:**
* Stable **facial/head features** for the same subject across images (eyes, nose, mouth, facial structure, hairstyle/color).
* Consistent **body shape** and **skin tone** for each individual across images.
* **Clothing/accessories** stability unless the instruction implies changes; otherwise do not penalize expected edits.
* For **multi-person scenes**, ensure each person maintains a consistent identity mapping across images (no A/B swapping).
**Ignore** differences in pose, composition, viewpoint, background, or lighting that do not affect identity.
**Scoring should be strict** — do not award high scores unless identity consistency is clear across all images.
Editing instruction: instruction
}

\subsection{User Study}
\label{sec:supp_user_study}

We invite 50 participants, who are familiar with image and video generative models, to engage in our evaluation on storyboard generation. We curate twenty evaluation samples by first searching some high quality human photos from website, then manually craft some storyboard caption based on them. 
For fairness, we include reference subjects spanning three racial groups (black, white and yellow) and two genders (female and male), and we vary prompts from simple to complex. Each testing sample provides one or two reference characters and requests generation of two to four storyboard images.

\begin{figure*}[t]
    \begin{center}
 \includegraphics[width=6.9in]{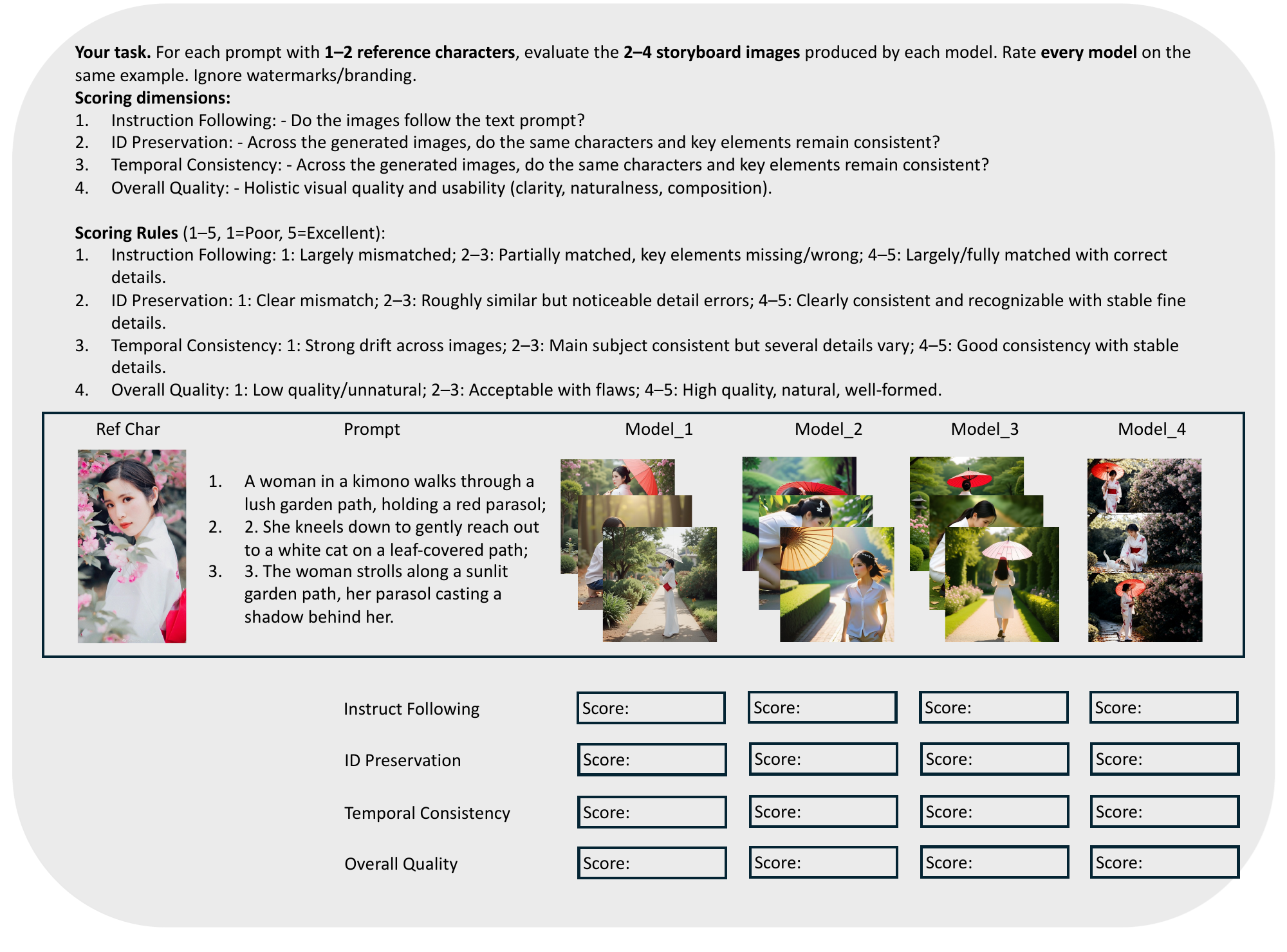}
\end{center}
    \captionof{figure}{User study template.}
\label{fig:user_study_template}
\end{figure*}

We request participants to rate for all results in the same sample. The rating system follows four criteria scored on a 5-point Likert scale (1=Poor, 5=Excellent): (i) \emph{Instruction Following}—whether the images follow the prompt; (ii) \emph{ID Preservation}—consistency with the reference character(s), emphasizing facial and fine attributes; (iii) \emph{Temporal Consistency}—whether the same character remains consistent across the generated panels; and (iv) \emph{Overall Quality}—a holistic judgment beyond adherence and consistency. For each sample, all competing models are rated by the same participant to reduce between-rater variance, and model identities are anonymized and presentation order is randomized. We then report scores for each metric based on the mean rating. We provide a showcase of our rating system in \cref{fig:user_study_template}. 

For a fair comparison, we use each model’s recommended inference settings. Specifically: StoryDiffusion at 768×768, classifier-free guidance (CFG)=5.0, 50 inference steps; UNO at 768×768, CFG=4.0, 25 steps; and OmniGen2 at 1024×1024, CFG=5.0, 50 steps. Our model follows setting as 1024x640 resolution, CFG=5.0, 50 steps. All experiments are conducted based on a random seed. Note that for a single sample, other models should be inferred several times with the same seed; iMontage uses one seed, outputting many results for one inference.

\begin{figure*}[t]
  \centering
  \includegraphics[width=\textwidth,height=0.9\textheight,keepaspectratio]{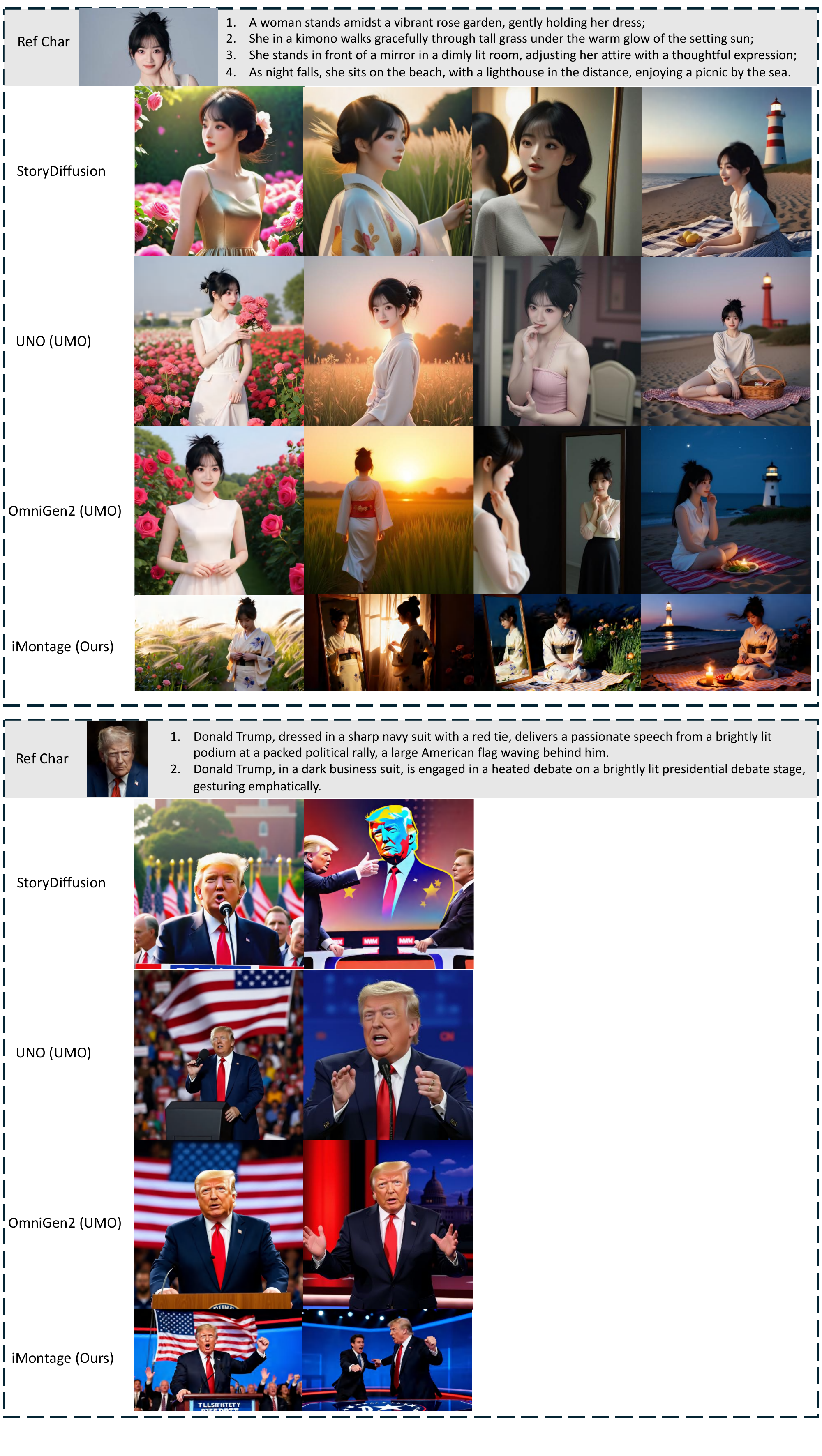}
  \caption{User study comparison visualization results. Zoom in to see more details.}
  \label{fig:user_study_vis_1}
\end{figure*}

\begin{figure*}[t]
  \centering
  \includegraphics[width=\textwidth,height=0.9\textheight,keepaspectratio]{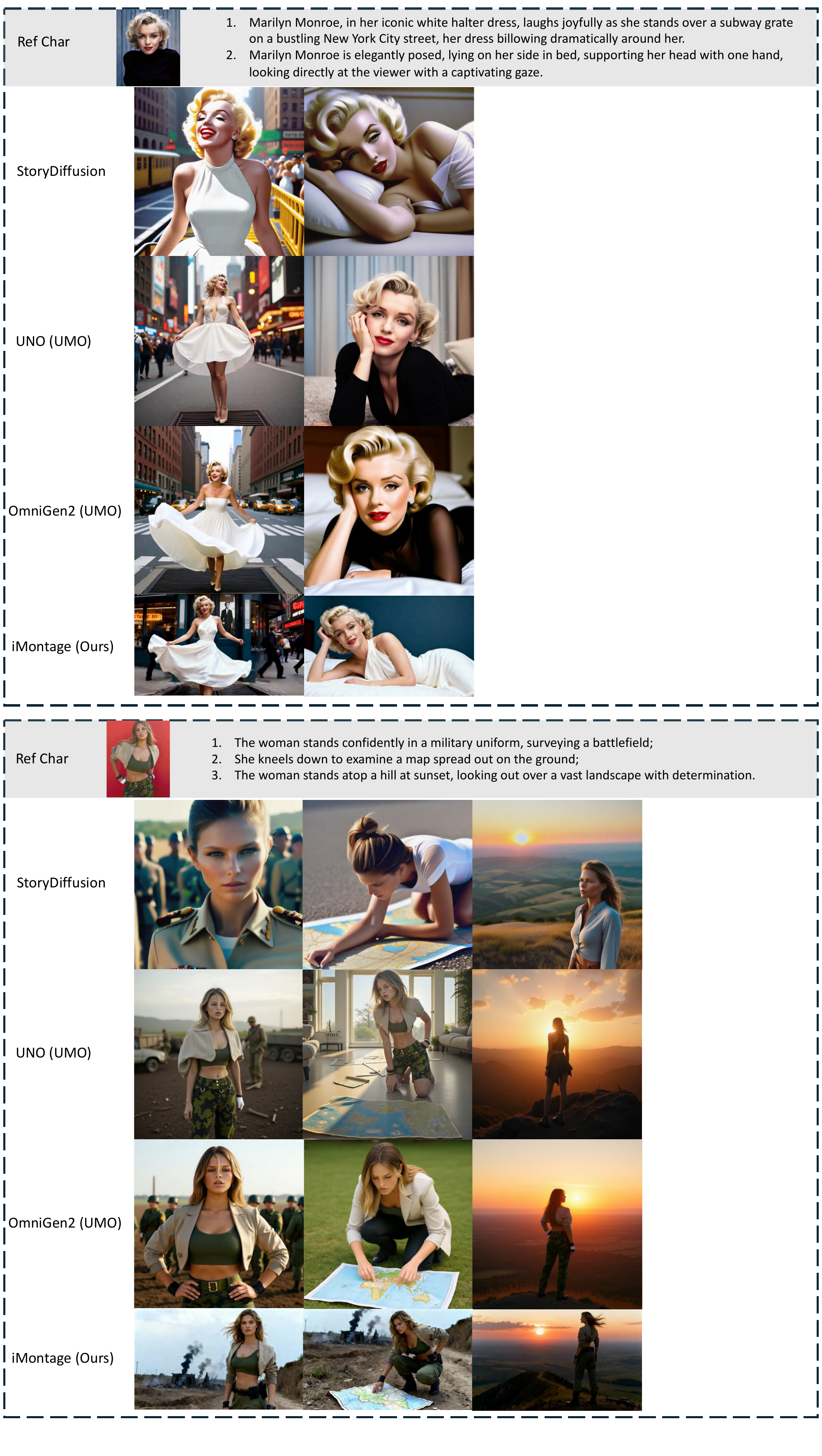}
  \caption{User study comparison visualization results. Zoom in to see more details.}
  \label{fig:user_study_vis_2}
\end{figure*}

\begin{figure*}[t]
  \centering
  \includegraphics[width=\textwidth,height=0.9\textheight,keepaspectratio]{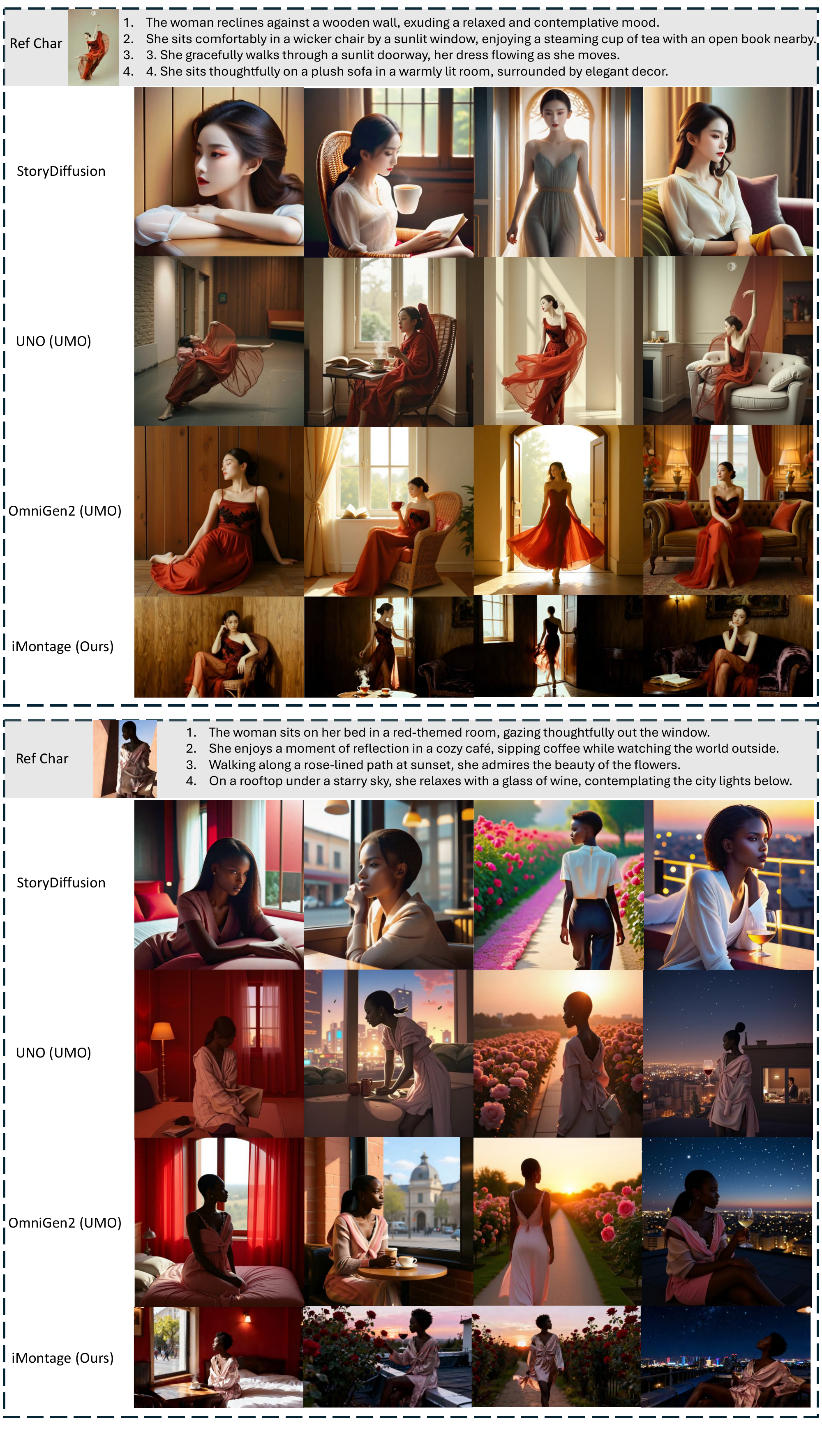}
  \caption{User study comparison visualization results. Zoom in to see more details.}
  \label{fig:user_study_vis_3}
\end{figure*}

\begin{figure*}[t]
  \centering
  \includegraphics[width=\textwidth,height=0.9\textheight,keepaspectratio]{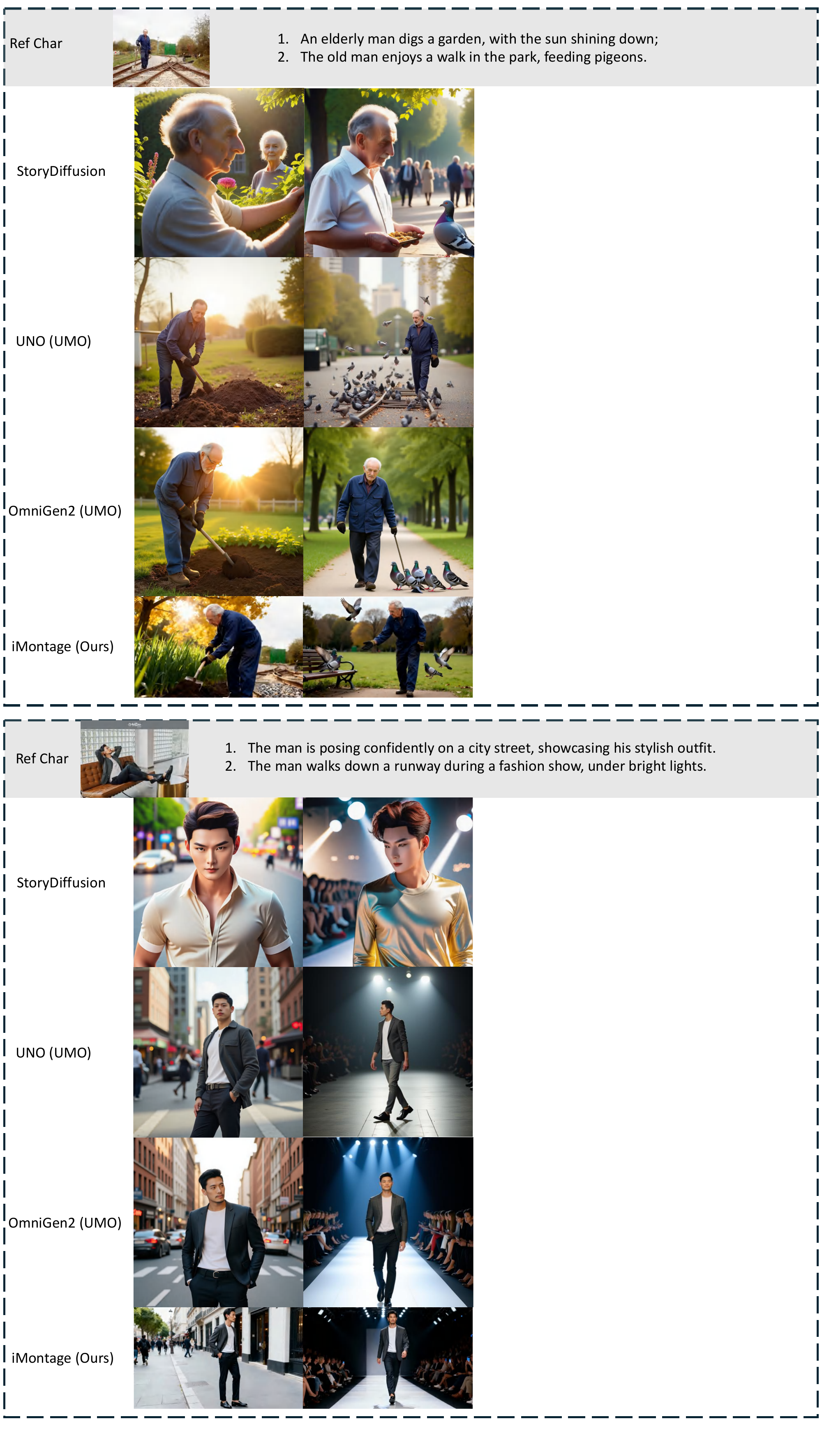}
  \caption{User study comparison visualization results. Zoom in to see more details.}
  \label{fig:user_study_vis_4}
\end{figure*}

\begin{figure*}[t]
  \centering
  \includegraphics[width=\textwidth,height=0.9\textheight,keepaspectratio]{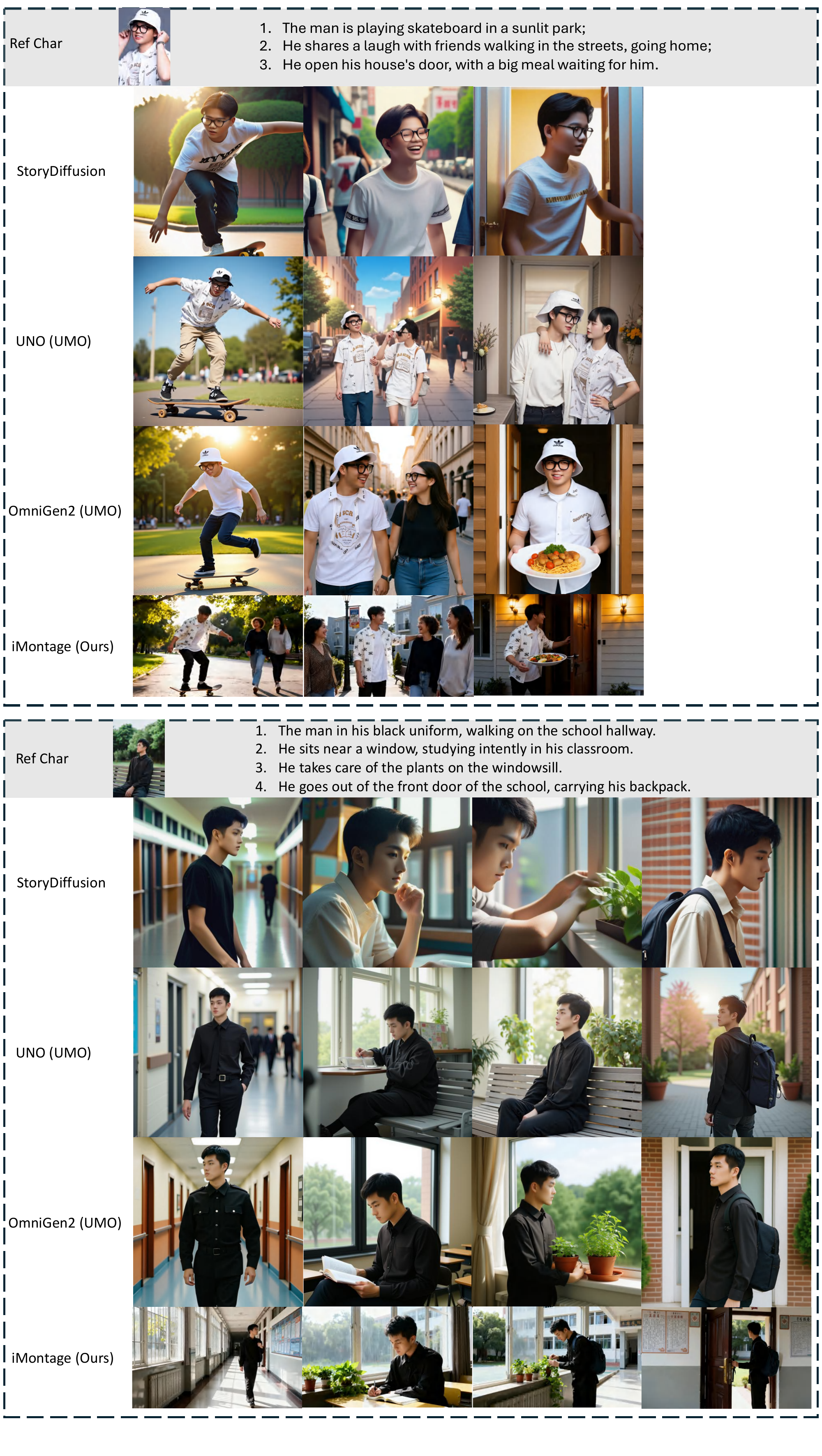}
  \caption{User study comparison visualization results. Zoom in to see more details.}
  \label{fig:user_study_vis_5}
\end{figure*}

\begin{figure*}[t]
  \centering
  \includegraphics[width=\textwidth,height=0.9\textheight,keepaspectratio]{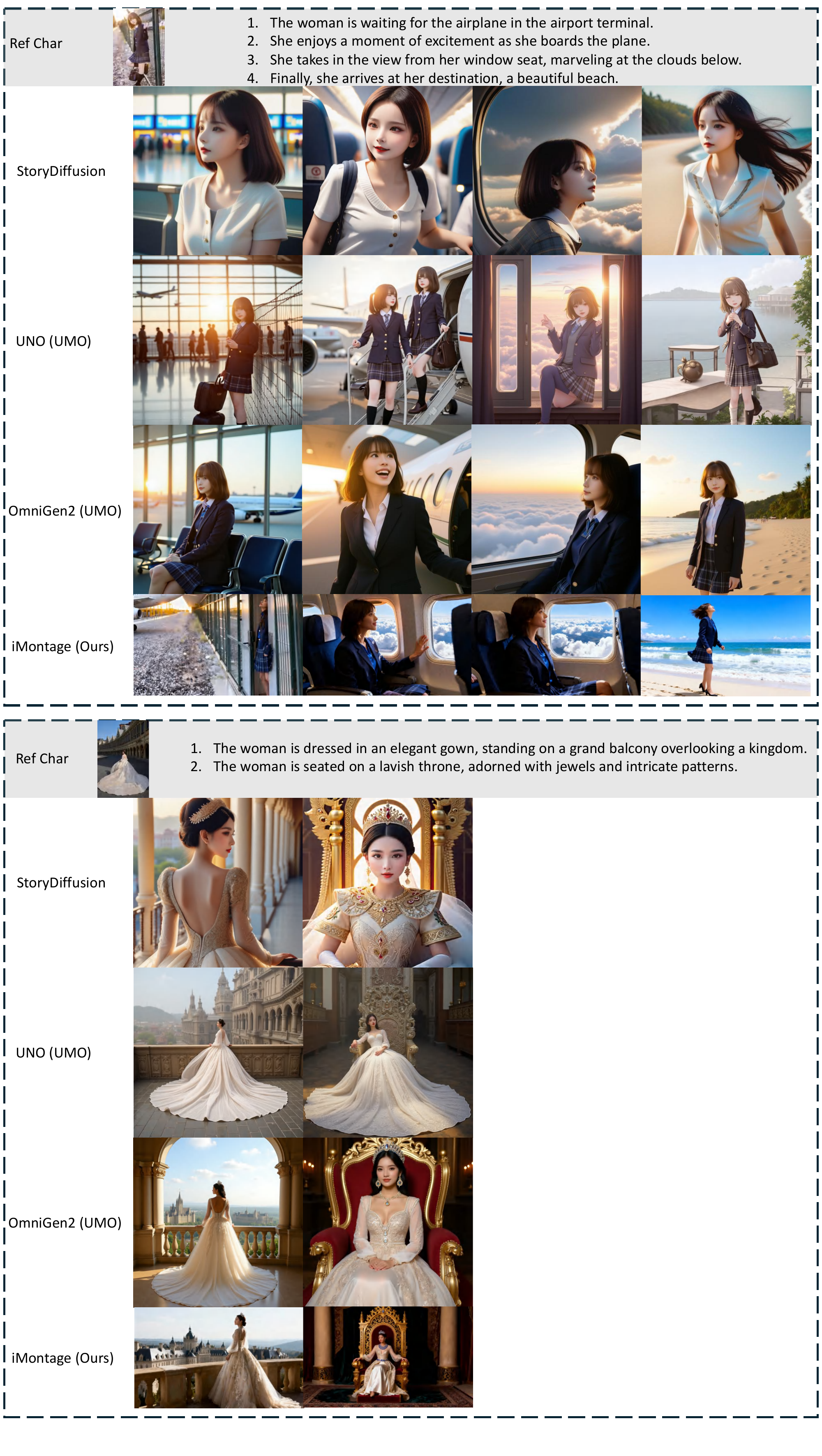}
  \caption{User study comparison visualization results. Zoom in to see more details.}
  \label{fig:user_study_vis_6}
\end{figure*}

\begin{figure*}[t]
  \centering
  \includegraphics[width=\textwidth,height=0.9\textheight,keepaspectratio]{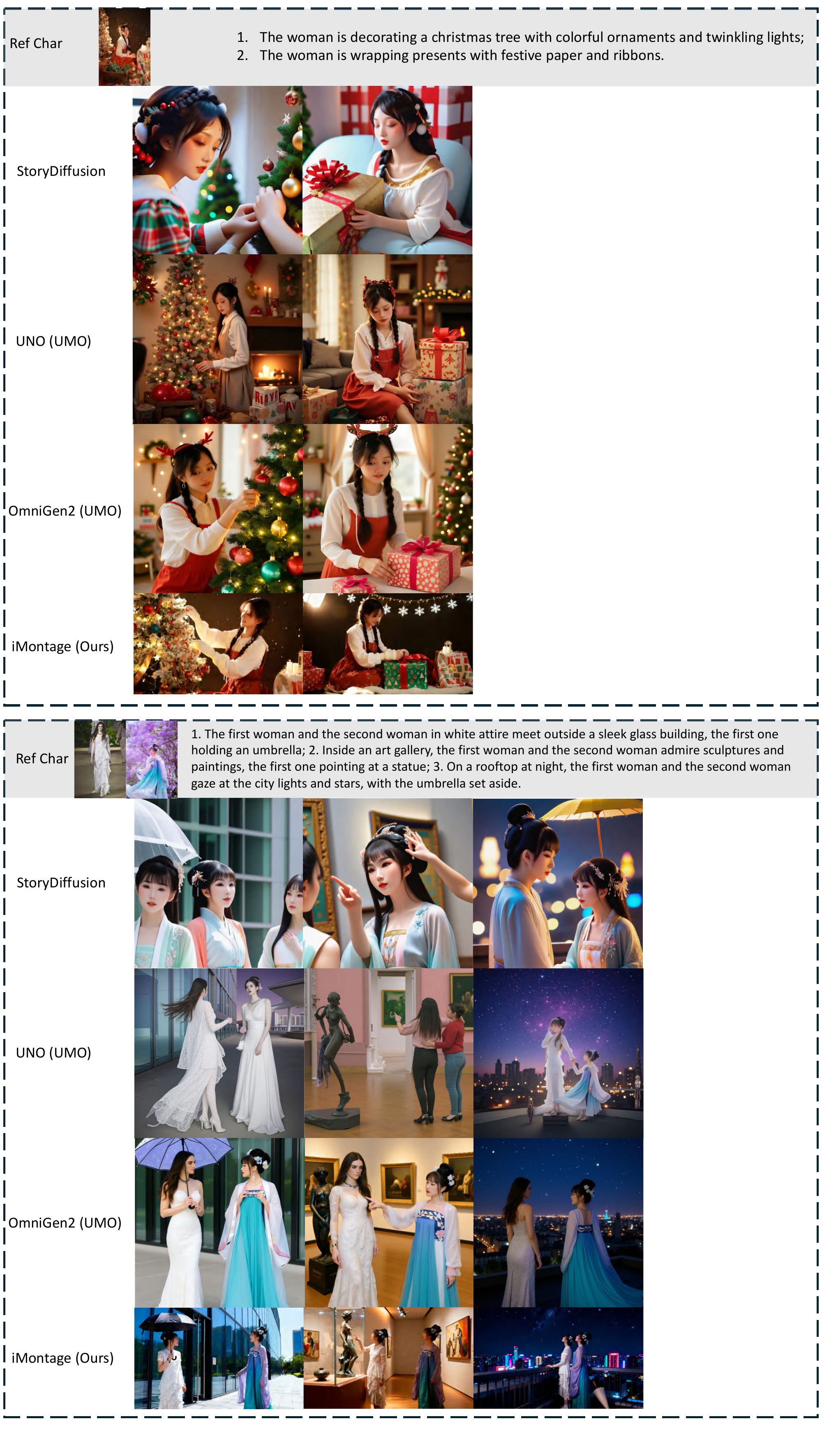}
  \caption{User study comparison visualization results. Zoom in to see more details.}
  \label{fig:user_study_vis_7}
\end{figure*}

We present the visualization results of evaluation from \cref{fig:user_study_vis_1} to \cref{fig:user_study_vis_7}.

\subsection{Training Scheme Ablation}
\label{sec:supp_ablation}
We ablate three scheduling strategies for SFT: \emph{FlatMix} (all tasks jointly), \emph{StageMix} (grouped by task type), and \emph{CocktailMix} (difficulty-ordered curriculum). We begin with \emph{FlatMix} and then transition to difficulty-aware scheduling.

\noindent \textbf{Task difficulty gap.}
Under a shared setup (data, optimizer, steps), we observe a clear difficulty spread across tasks: the easiest task, \emph{multi-editing}, and the hardest task, \emph{storyboard generation}, differ by roughly 
0.2 in training loss. This gap motivates difficulty-aware mixing.

\noindent \textbf{StageMix vs.\ CocktailMix.}
We train \emph{StageMix} with the same protocol used for our Stage~2 and Stage~3 runs and compare it head-to-head with \emph{CocktailMix}. On OmniContext\citep{wu2025omnigen2}, \emph{StageMix} underperforms by 12.6\% relative to \emph{CocktailMix}. Other tasks all have worser visualization results for \emph{StageMix}. These observations indicate that difficulty-ordered mixing yields better optimization stability and stronger generalization, especially on the harder tasks.

\section{More Discussion}

\subsection{Concurrent Works}
\label{sec:supp_concurrent_works}
Though we are not the first unified image generation model developed upon video models\citep{chen2025unireal, lin2025realgeneral}, we consider iMontage as the first practical many-to-many system for open-source community. Likewise, two very recent efforts build image capabilities on top of video backbones. ChronoEdit\citep{wu2025chronoedit} treats the input and edited outputs as the first and last frames of a short “video” and jointly denoises them with temporal-reasoning tokens, leveraging a pretrained video generator to improve physical plausibility and temporal coherence in edits. UniVid\citep{chen2025univid} explores a complementary route: it adapts a pretrained video DiT with lightweight SFT to a broad suite of vision tasks—both understanding and generation—by casting tasks as “visual sentences,” thereby avoiding task-specific architectural changes and generalizing across modalities and data sources. 

Our model focuses on another area, narrowing the gap between image and video generation by casting image synthesis as a unified many-to-many problem. We view this as a practical technical pathway and plan to extend it into a more capable, fully unified system.

\subsection{Observed Failure Case}
\label{sec:supp_failure_case}

\begin{figure*}[t]
  \centering
  \includegraphics[width=\textwidth,height=\textheight,keepaspectratio]{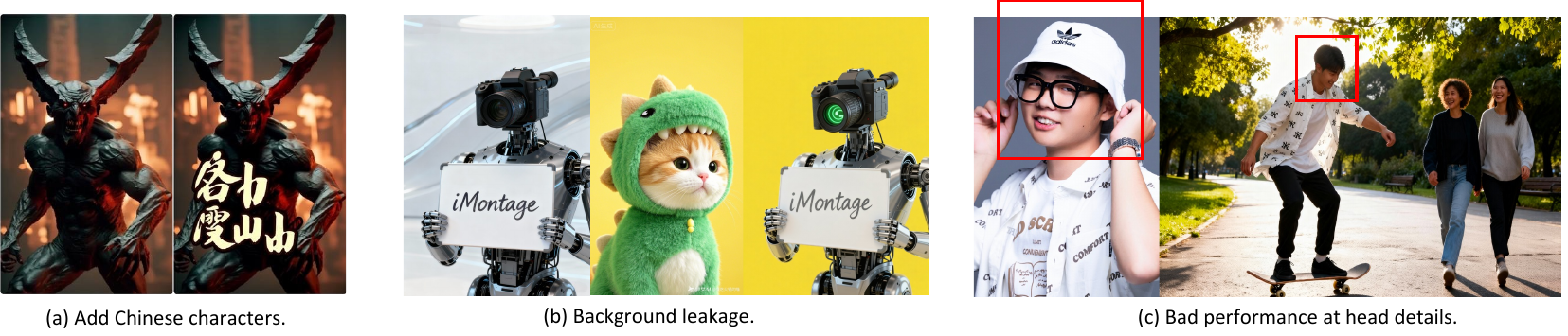}
  \caption{Representative failure case for a certain task. Zoom in to see more details.}
  \label{fig:failure_case}
\end{figure*}

Our model still exhibits failure cases on certain tasks, as illustrated in \cref{fig:failure_case}.
For \textbf{image editing}, the most salient issue is near-zero ability to render Chinese characters (\cref{fig:failure_case}a), largely inherited from the base backbone HunyuanVideo \citep{kong2024hunyuanvideo}, which lacks robust text-rendering supervision.
For \textbf{SRef}, our training data are distilled from other models \citep{wu2025uso, xing2024csgo}, which is suboptimal; we observe occasional background leakage, which is a known challenge in style-reference transfer.
Finally, we note a \textbf{head-detail mismatch} in some generations. This limitation stems from data constraints—namely, insufficient training coverage of diverse, high-detail head/face depictions. Two complementary remedies are promising: (i) adopt human-centric identity modules by injecting face embeddings \citep{ye2023ip, wang2024instantid, xu2025withanyone}; and (ii) expand coverage of high-quality, head-focused data to strengthen fine-grained facial detail preservation.


\begin{figure*}[t]
  \centering
  \includegraphics[width=\textwidth,height=\textheight,keepaspectratio]{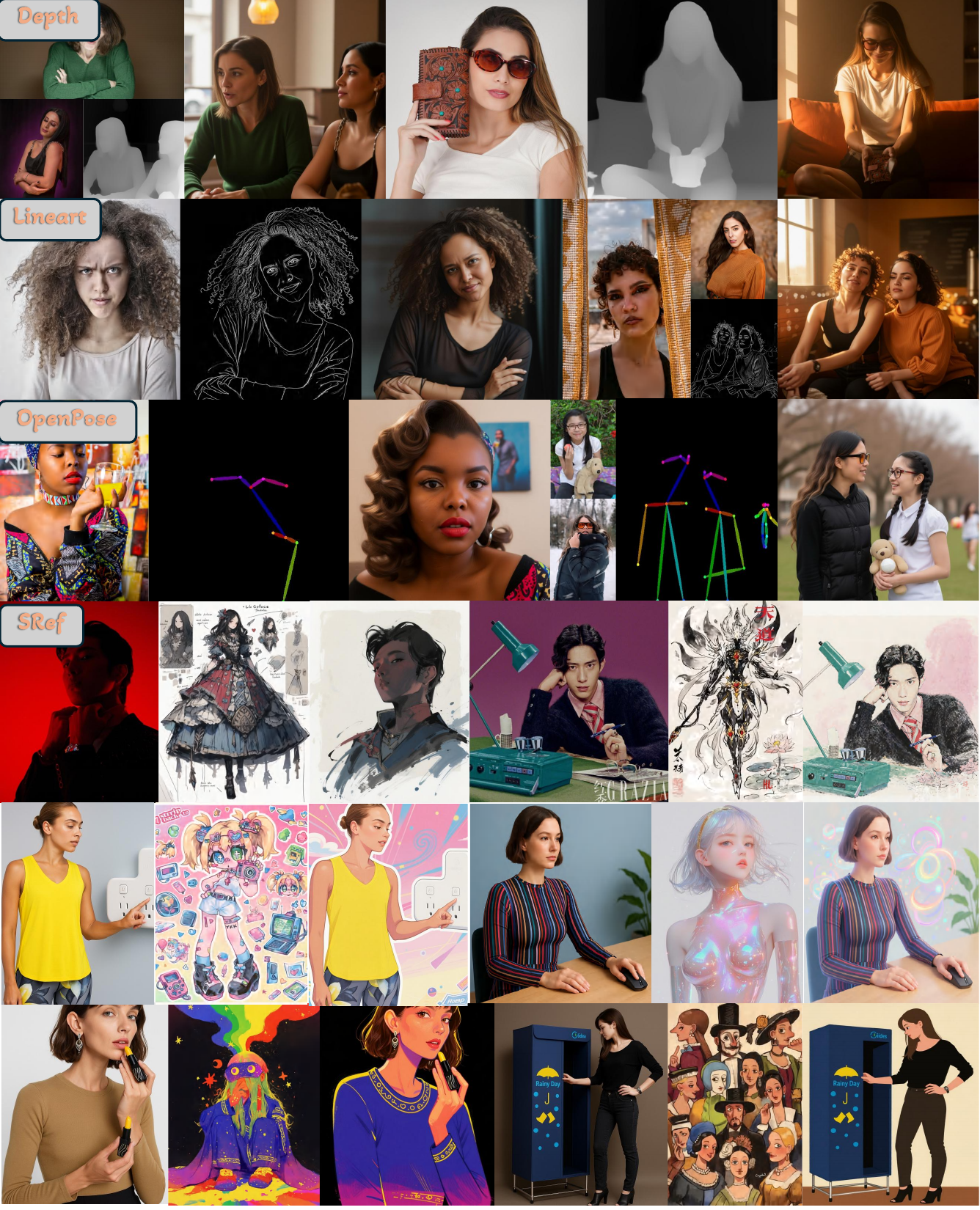}
  \caption{Visualization results for conditioned CRef and SRef. Zoom in to see more details.}
  \label{fig:conditioned_cref_sref_vis}
\end{figure*}

\end{document}